\documentclass[letterpaper, 10 pt, conference]{ieeeconf}
\pdfoutput=1
\usepackage{hyperref}
\usepackage{algorithm}
\usepackage{algpseudocode}
\usepackage{subfig}
\usepackage{graphicx}
\usepackage{algorithmicx}
 
\usepackage{amsthm}
\usepackage{tabularx}
\usepackage{cite}
\usepackage{amsmath,amssymb,amsfonts}
\usepackage{textcomp}
\usepackage{xspace}
\usepackage{makecell}
\usepackage{booktabs,siunitx}
\usepackage{diagbox}
\usepackage{array, multirow}

\begin{document}

\title{Nonlinear Correct and Smooth for Semi-Supervised Learning}
\author{
Yuanhang Shao$^{1}~~~$
Xiuwen Liu$^{1}~~~$
\smallskip 
\vspace{0.25em} \\
$^1$\small{Department of Computer Science, Florida State University}
\\
\small{\{shao, liux\}@cs.fsu.edu}
}

\maketitle

\begin{abstract}
Graph-based semi-supervised learning (GSSL) has been used successfully in various applications. Existing methods leverage the graph structure and labeled samples for classification. Label Propagation (LP) and Graph Neural Networks (GNNs) both iteratively pass messages on graphs, where LP propagates node labels through edges and GNN aggregates node features from the neighborhood. Recently, combining LP and GNN has led to improved performance. However, utilizing labels and features jointly in higher-order graphs has not been explored. Therefore, we propose Nonlinear Correct and Smooth (NLCS), which improves the existing post-processing approach by incorporating non-linearity and higher-order representation into the residual propagation to handle intricate node relationships effectively. Systematic evaluations show that our method achieves remarkable average improvements of 13.71\% over base prediction and 2.16\% over the state-of-the-art post-processing method on six commonly used datasets. Comparisons and analyses show our method effectively utilizes labels and features jointly in higher-order graphs to resolve challenging graph relationships.
\end{abstract}

\section{Introduction}
Graphs are indispensable in the realm of big data for representing data generated from non-Euclidean domains, as they consist of interconnected nodes carrying complex relationships and dependencies between entities. Recently, the growing variety of graph-based real-world applications, such as social networks \cite{butts2008social}, citation networks \cite{dawson2014current}, and recommendation systems \cite{adeniyi2016automated}, have heightened the necessity for effective models to capture the underlying intricate relationships. Graph-based semi-supervised learning (GSSL) targets node classification with given a subset of labeled nodes. A hallmark of real-world graphs is homophily \cite{zhu2020beyond}, where neighbor nodes tend to share the same labels or exhibit similar features, and almost all existing GSSL methods exploit that. However, most studies in GSSL have primarily focused on graph representations (i.e., nodes and edges) using either labels or features \cite{zhou2003learning, huang2020combining, verma2021graphmix, ding2022meta}. Effectively and jointly leveraging labels, features, and higher-order representations to maximize the generalization performance of GSSL remains a challenge.

Traditional approaches in GSSL have primarily focused on utilizing homophily within labels to develop smoothing techniques, ranging from early methods like random walks \cite{szummer2001partially} to more recent label propagation (LP) \cite{zhou2003learning}. The standard LP algorithm spreads the known labels to adjacent neighbors through edges. Further research has highlighted the improvement gained from incorporating higher-order representations (such as triangles) and non-linearity, which allows to capture higher-order triangles more accurately \cite{tudisco2021nonlinear}. On the other hand, with the increasing availability of data and the popularity of deep learning models, Graph Neural Networks (GNNs) have emerged as powerful models focusing on learning expressive node representations and smoothing node features \cite{scarselli2008graph}. Unlike fully connected networks treating inputs as an ordered list, GNNs generalize neural networks to effectively capture the complex relationships of graphs through aggregating neighbor features and smoothing to reduce noise \cite{li2018deeper, feng2020graph, verma2021graphmix}. Nonetheless, GNNs are sensitive to hyperparameters and lack meaningful representation for the smoothed output \cite{feng2020graph, li2018deeper,oono2019graph, li2019deepgcns}. Recent studies have shown that combining label spreading with deep learning approaches can yield competitive performance \cite{huang2020combining, wang2021combining, ding2022meta}. These works emphasize the benefits of using labels as auxiliary knowledge rather than solely relying on features. However, it is worth noting that the joint utilization of labels and features in higher-order graphs remains unexplored in the current literature.

To address the aforementioned challenges, we propose Nonlinear Correct and Smooth (NLCS), a framework that leverages both labels and features in a joint manner. NLCS starts by utilizing a neural network model to learn node representations as base prediction and then propagates residuals to correct labels using nonlinear and higher-order graph relationships. Specifically, NLCS is inspired by Correct and Smooth (C\&S) \cite{huang2020combining}, where we enhance their post-processing by incorporating non-linearity and higher-order representation from Nonlinear Higher-order Label Spreading (NHOLS) \cite{tudisco2021nonlinear} into the residual propagation. The introduction of higher-order representations improves discrimination between classes and provides stronger modeling flexibility, resulting in further improvements in correction and smoothing. Through extensive experimentation, the proposed method shows an average improvement of 13.71\% over base prediction and 2.16\% compared to C\&S, thereby validating the effectiveness of our joint utilization of labels and features on higher-order graphs.

\section{Related work}
In recent years, significant progress in deep learning and graph analysis has led to various new and improved methods for solving GSSL. Here we provide a brief review to investigate several common approaches utilizing labels and features in existing GNNs, LP, and combined methods.

\textbf{Graph Neural Networks} (GNNs) are deep neural networks designed to operate on graph-structured data utilizing the inherent structure and relationships within graphs for solving various graph-based tasks ranging from social networks to citation networks \cite{verma2021graphmix, ding2022meta}. GNNs generally aggregate information from neighbors to generate new node representations, which can express graph structure effectively with multi-hop aggregation. Previous works of GNNs utilize smoothed node features to achieve promising performance in GSSL \cite{battaglia2018relational, chong2020graph}. However, recent works show that the GNN-based models intrinsically suffer from over-smoothing and information loss, which resulting indistinguishable representations of nodes across different labels \cite{feng2020graph, thekumparampil2018attention, li2018deeper,oono2019graph}. 

To elaborate on the inherent limitations of GNNs, recent studies focus on understanding insights of GNNs and techniques to alleviate over-smoothing, such as skip connection, graph normalization, random dropping \cite{li2018deeper, feng2020graph, li2019deepgcns, chen2020simple, zhou2021understanding, huang2020tackling}. \cite{li2018deeper} highlight that the graph convolutional network (GCN) is a special form of Laplacian smoothing, which can smooth the noise of node features. However, this smoothing process can lead to less distinguishable features among different clusters. In the attempt to elucidate the effectiveness and limits of GNNs, \cite{chen2022bag} show that deeper GNNs are more susceptible to over-smoothing. Even though the initial and jumping connection improves the performance of deep GNNs, the normalization techniques exhibit inconsistent performance on different GNN backbones and become unstable with deeper networks; and the random dropping mitigates performance degradation but fails to improve deep GNNs consistently. Furthermore, \cite{oono2019graph} emphasize that a deeper network and the introduction of non-lineality can aggravate the over-smoothing issue. Therefore, GNNs suffer from the inherent trade-off between smoothing noise and making node features distinguishable, and the effectiveness of initial and jumping connections suggests the importance of preserving the original feature vectors and their correlation with labels. Intuitively, exploring alternative methods that leverage the labeled data and mitigate information loss caused by aggregation is worthwhile. 

\textbf{Label propagation} (LP) is a well-known technique for spreading known labels to unlabeled nodes based on the homophily principle. \cite{zhou2003learning} proposed the standard LP algorithm, which is an iterative propagation function spreading labels via the structure of graphs. Moreover, \cite{tudisco2021nonlinear} proposed a nonlinear label spreading function, incorporating the higher-order representations of graphs, known as Nonlinear Higher-order Label Spreading (NHOLS). Notably, the NHOLS outperforms the GraphSAGE, a two-layer GNN that incorporates concatenation-based skip connections to alleviate over-smoothing, while NHOLS achieves superior performance by leveraging nonlinear mixing functions and spreading information over hyperedges. These straightforward approaches indicate that the propagation technique can effectively leverage the advantages of known labels and nonlinearity. 

\textbf{Combined methods} combine GNNs with regular neural networks or LP to improve prediction accuracy, which has gained increasing interest in recent years  \cite{huang2020combining, wang2021combining, jia2022unifying, ding2022meta}.  \cite{ding2022meta} propose a framework (known as Meta-PN) that considers propagated label predictions as pseudo labels and utilizes multi-layer perceptron (MLP) to predict the labels based on pseudo labels. Their label propagator is an adaptive version involving trainable parameters to adjust the influence of $k$-hop neighborhood. On the other hand, \cite{huang2020combining} proposes a simple post-process step (C\&S) based on residual propagation and smoothing to improve the base prediction of deep neural networks, including MLP and GNNs. This approach provides a new perspective for solving semi-supervised learning and opens up new avenues for exploring the combination of LP and deep neural networks. Similarly, \cite{wang2021combining} introduce the concept of influences and studies the influences of features and labels in label propagation algorithm (LPA) and graph convolutional neural networks (GCN). They proposed a unified model to make the edges trainable based on label influence, as nodes with the same label should have strong connections. Another notable work, \cite{jia2022unifying} propose a Markov random field model (MRF) for node attributes generation, such that the label propagation, linearized graph convolutional network, and their combination can be derived as conditional expectations under the MRF model. Overall, label propagation proves to be a powerful method and can be flexibly applied to labels, features, residuals, and smoothing. These combined methods show more promise in utilizing labels and features jointly to solve semi-supervised learning tasks. However, none of these combined methods have explored the potential of higher-order representation. By conducting nonlinear higher-order propagation on top of base predictions, our approach NLCS achieves superior performance on GSSL. 

\section{Preliminary}
\label{preliminary}
\subsection{Label spreading}
\label{ls}
 The most well-known label spreading (LS) algorithm was proposed by \cite{zhou2003learning}, and we discuss their approach as it provides the foundation for our method and allows us to inherit some of the notations. Assuming an undirected and weighted graph $G=(V, E, \omega)$, where $V=\{1, 2, \dots, n\}$ represents the set of nodes with their corresponding feature vector represented as $X \in \mathbb{R}^{n \times p}$, $E\subseteq V \times V$ represents the edge set, and each edge $e \in E$ is associated with a positive weight $\omega (ij) > 0$. Let $A$ be the adjacency matrix of $G$, such that $A_{ij} = \omega (ij), ij \in E$, and $A_{ij}=0$ otherwise; $D_G = Diag(d_1, d_2, \dots, d_n)$ be the diagonal degree matrix; and $S = D_G^{-\frac{1}{2}}AD_G^{-\frac{1}{2}}$ be the normalized adjacency matrix. The task is to predict a subset of unlabeled nodes $U \subset V$ based on a small and disjoint subset of labeled nodes $L \subset V$ using a set of labels $\{1,2,\dots,C\}$, where $U + L = V$. Therefore, a one-hot-encoding matrix $Y \in \mathbb{R} ^{n\times c}$ is used to present the label of nodes, where $c$ is the number of labels and $Y_{ij} = 1$ if $i \in L$ for label $j$, and 0 otherwise. The standard LP iteratively updates labels using the propagation function:
\begin{equation}
\label{eq:lp}
    F^{(t+1)} = \alpha SF^{(t)}+(1-\alpha)Y,
\end{equation}
where $0 < \alpha < 1$, and $F^{(0)} \in \mathbb{R}^{n \times c}$. This equation converges to the solution of the linear system $(I-\alpha S)F^*=(1-\alpha)Y$ after iterations, and the approximate optimal solution is denoted as $\tilde{F}^*$. 

\subsection{Nonlinear higher-order label spreading}
\label{nhols}
In unsupervised and semi-supervised learning, higher-order relations are commonly used because the edges of hypergraphs are non-empty subsets of the vertices and are more flexible and effective in modeling the relationships of a graph. By utilizing the higher-order representation of graphs, the performance can be improved in various approaches, such as label spreading \cite{eswaran2020higher, tudisco2021nonlinear}. Our work adopts some of the notations from the NHOLS \cite{tudisco2021nonlinear}. Given a 3-regular hypergraph $H=(V, \varepsilon, \tau)$ to present the weighted graph $G$, where $\varepsilon \in V \times V \times V$ and $\tau$ is the weight of hyperedges, such that $\tau_{ijk}>0$ and hyperedges are a set of triangles of $G$. Therefore, the third-order adjacency tensor is defined as $\mathcal{A}$, where $\mathcal{A}_{ijk}=\tau_{ijk}$ and 0 otherwise. Furthermore, we can define the diagonal matrix of the hypergraph as $D_H=Diag(\delta_1, \dots, \delta_n)$. The nonlinear mixing function is denoted as $\sigma: \mathbb{R}^2 \to \mathbb{R}$, which the tensor mapping is defined entrywise for the third-order adjacency tensor $\mathcal{A}$:
\begin{equation}
    \mathcal{A} \sigma(f)_i = \sum_{jk} \mathcal{A}_{ijk} \sigma(f_i,f_k).
\end{equation}
Then, this tensor mapping function is represented in matrix format to reflect the nonlinear part of NHOLS:
\begin{equation}
    \mathcal{S}(f) = D_H^{-\frac{1}{2}} \mathcal{A}\sigma(D_H^{-\frac{1}{2}}f).
\end{equation}
At last, the NHOLS iteration is defined as follows:
\begin{equation}
    G^{(t)} = \alpha \mathcal{S}(F^{(t)})+\beta SF^{(t)} + (1-\alpha-\beta)Y, F^{(t+1)}=G^{(t)}/\varphi (G^{(t)}),
\end{equation}
where $\varphi (G^{(t)})$ is the normalization at iterations defined as:
\begin{equation}
\label{eq:phi_norm}
    \varphi (f) = \frac{1}{2} \sqrt{\sum_{ij}B_{ij}\sigma(f_i/\sqrt{\delta_i},f_j/\sqrt{\delta_j})^2},
\end{equation} 
and $B$ is the matrix $B_{ij}=\sum_{k}\mathcal{A}_{kij}$ derived from $\mathcal{A}$. In the next section, we inherit the same graph and hypergraph notation as defined in previous works and propose our approach.

\begin{figure}[t]
    \centering
    \includegraphics[width=0.6\linewidth]{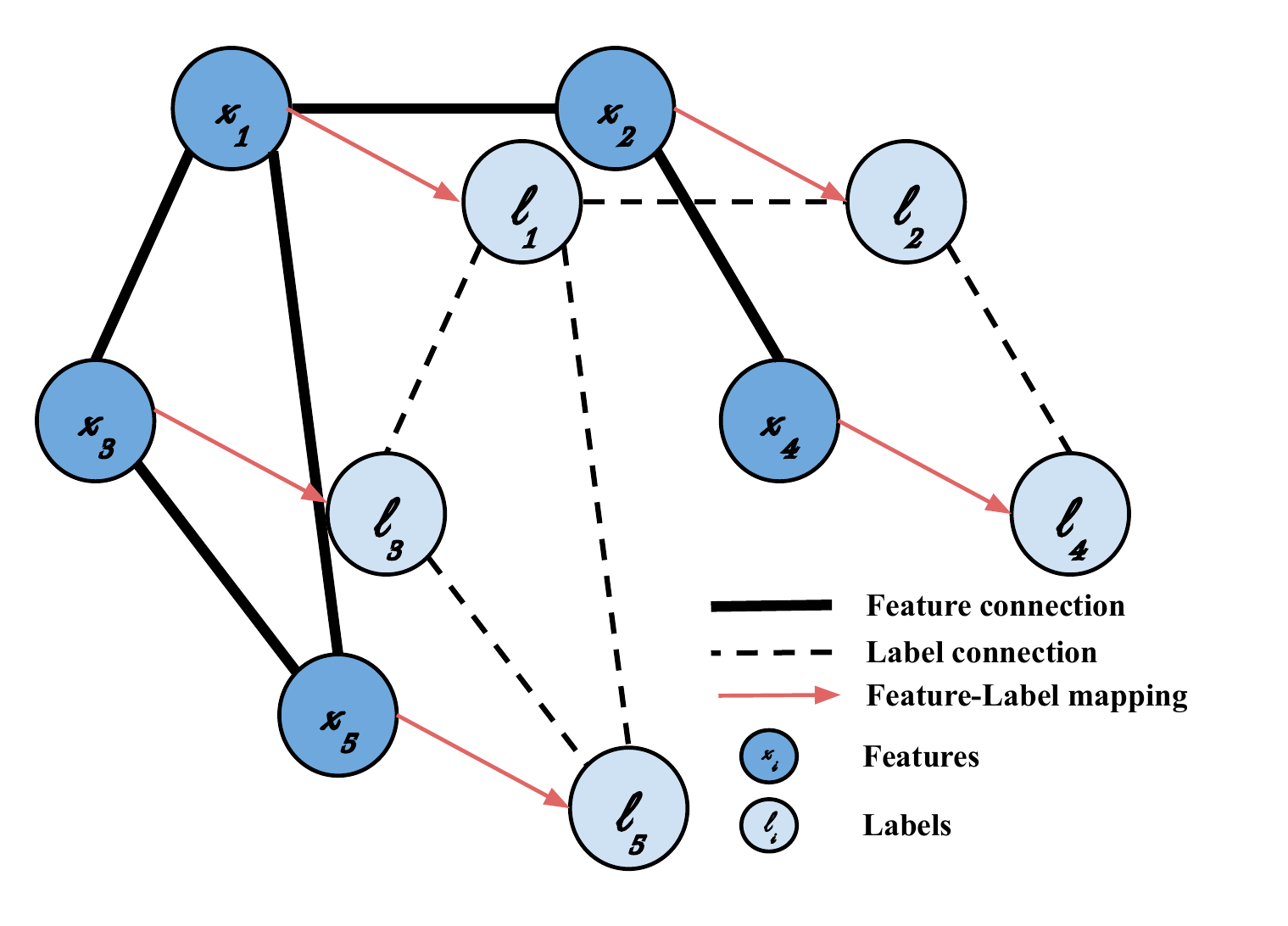}
    \hfill   
  \caption{Overview of NLCS framework. }
  \label{fig-nlcs-frame} 
\end{figure}

\section{NLCS Framework}

\label{frame-nlcs}
First of all, let $G=(V, E)$ be an undirected graph, where $V$ represents the set of $n$ vertices labeled by $l=\{1, \dots, c\}$ and $E$ is the set of $m$ edges. As illustrated in Figure \ref{fig-nlcs-frame}, vertices can be represented by their feature vectors $\textbf{\textit{x}}$, which are mapped to labels $\textbf{\textit{l}}$ using neural networks; then, the residual of neighbors is assumed to be similar in base prediction, such that the error can propagate through the edges of identical graph structure connected with dashed lines. The first mapping step can be modeled as Markov Random Field (MRF) prior to independent observations \cite{jia2022unifying}, where the model has direct links from $\textbf{\textit{l}}$ to $\textbf{\textit{x}}$, indicating an independent relationship denoted by $\textit{Pr}(\textbf{\textit{x}}|\textbf{\textit{l}})$ and estimated by neural networks. The real-world data is usually sparsely connected such that each feature only matches with one label, and each node is connected to only a few others. In this scenario, the MRF model should be conditionally independent of all of the other variables given its neighbors, and describes the joint probability of the variables as a product of potential function as follows:
\begin{equation}
\label{eq-mrf}
    \textit{Pr}(\textbf{\textit{l}})= \frac{1}{Z}\prod_{j=1}^{J}\phi_j[\textbf{\textit{l}}_{\mathcal{C}_j}],
\end{equation}
where $\phi[\bullet]$ is the potential function depending on the subset of labels, which are neighbors in our case. The term $Z$ is a normalizing constant that ensures the result is a valid probability distribution.   

After modeling the problem as MRF, we can proceed to solve the problem with compound Bayesian decision theory using Eq. \ref{eq-mrf} derived from the MRF model. Our objective is to predict the overall graph including unlabeled nodes $\textbf{w}=\{w_1,\dots, w_N\}$ based on the observation $\textbf{x}=\{x_1,\dots, x_N\}$ consisting of discrete variables. We can compute the posterior probability over the unlabeled nodes using Bayes' rule within MRF:
\begin{equation}
    \textit{Pr}(l_{1\dots N}|x_{1\dots N}) = \frac{\prod_{n=1}^{N}\textit{Pr}(x_n|l_n)\textit{Pr}(w_{1\dots N})}{\textit{Pr}(x_{1\dots N})},
\end{equation}
where we assume that the conditional probability should be a product of individual terms associated with neighborhoods. Overall, we use a neural network model to learn the representation of nodes and make a base prediction on $\textit{Pr}(\textbf{\textit{x}}|\textbf{\textit{l}})$. Subsequently, we use residual correction to make joint decisions based on the base predictions of $n$ nodes.

\section{Nonlinear Correct and Smooth}
Our approach is a post-processing step, inspired by the concept of Correct and Smooth (C\&S) \cite{huang2020combining}, which aims to improve the prediction accuracy of simple learning models by conducting residual propagation through higher-order graphs. C\&S assumes that the errors of neighboring nodes are likely similar to those in the base prediction results \cite{huang2020combining}. However, the neighbors are more complex in real-world graphs and not necessarily surrounded by the same labels, which can mislead the prediction, such as in Figure \ref{fig-rice}. Therefore, we consider error spreading over nonlinear higher-order representation to handle the complex relationship within the graph. As mentioned, our approach inherits graphs and higher-order graphs notation defined in section \ref{preliminary}. First, we define an error matrix $E \in \mathbb{R}^{n \times c}$ based on the one-hot-encoding matrix $Y$, which is split into two parts:
\begin{equation}
    E_L = X_L - Y_L, E_U = 0,
\end{equation}
where $X_L$ is the base prediction of labeled nodes, and the error is the residual on labeled and 0 on unlabeled. Then, we propagate errors at node $i$ to neighboring nodes of $i$ which should increase the similar error to surrounding nodes. Considering the notation as mentioned in section~\ref{nhols}, we use the nonlinear mixing functions $\sigma:\mathbb{R}^2 \to \mathbb{R}$ to define the tensor mapping $\mathcal{A}\sigma: \mathbb{R}^{n\times c} \to \mathbb{R}^{n\times c}$ regarding to the error matrix entrywise as follows:
\begin{equation}
    \mathcal{A} \sigma(E)_i = \sum_{jk} \mathcal{A}_{ijk} \sigma(E_j,E_k).
\end{equation}
This tensor mapping aggregates errors from neighbors of higher-order graphs via a nonlinear mixing function, such as triangles which consider errors from two adjacency nodes of a triangle and its cosine of the angle. Then, the nonlinear mapping for the error matrix is denoted by:
\begin{equation}
    \mathcal{S}(E) = D_H^{-\frac{1}{2}} \mathcal{A}\sigma(D_H^{-\frac{1}{2}}E).
\end{equation}
At last, the residual propagation is iterated by:
\begin{equation}
\label{eq:nlc}
    E^{(t+1)} = \alpha \mathcal{S}(E^{(t)})+\beta SE^{(t)} + (1-\alpha-\beta)Y,
\end{equation}
where the initial vector $E_{(0)}=E$. This iteration function propagates the error in third-order adjacency representation or higher, in which the higher-order propagation is nonlinear. The error propagation is provable under a Gaussian assumption in regression problems \cite{jia2020residual}, and the errors in equation~\ref{eq:nlc} need normalization to adjust the scale of residual. We adopt the Autoscale \cite{huang2020combining}, in which the scale of the error is approximated with the average error of the labeled nodes. The L1-norm for labeled nodes is denoted as:
\begin{equation}
    \lambda = \frac{1}{|L|} \sum_{j\in L}||e_j||_1,
\end{equation}
where $e_j \in \mathbb{R}^c$ is the $j$th row of $E$. Eventually, the base prediction of unlabeled nodes is corrected by the error that is propagated through the higher-order nonlinear function as follows:
\begin{equation}
\label{eq:nlc'}
    X_{i}' =  X_i + \lambda \frac{\widehat{E}_i}{||\widehat{E}_i||_1}, i \in U,
\end{equation}
where $\widehat{E}_i$ is the approximated error of $i$th unlabeled node after the error propagation iteration in equation~\ref{eq:nlc}.

After the correction using propagated residual, our approach conducts smoothing based on corrected labels. Considering the diversity of graphs, the residual correction handles heterophily neighbors to provide good estimations. The corrected predictions of unknown labels are considered relatively accurate compared to base predictions. Therefore, smoothing can refine and improve the overall prediction accuracy by applying additional label propagation over the entire graph. This step utilizes the best predicted labels $G \in \mathbb{R}^{n\times c}$:
\begin{equation}
    G_L = Y_L, G_U = X_U',
\end{equation}
where the labeled nodes use the true label of training and the unlabeled nodes use the corrected prediction in equation~\ref{eq:nlc'}. Subsequently, the final prediction $\hat{Y}$ is approximated iteratively using the nonlinear higher-order label spreading function $G^{(t+1)} = (\alpha \mathcal{S}(G^{(t)})+\beta SG^{(t)} + (1-\alpha-\beta)Y)/\varphi (G^{(t)}),$ where $G^{(0)}=G$ and $\varphi$ denotes the normalization process described in equation~\ref{eq:phi_norm}. Finally, for each node $i \in U$, the prediction of the label is assigned as the one with the highest probability among the classes denoted as $argmax_{j\in \{1,2,\dots,C\}}\hat{Y}_{ij}$.

\begin{table*}[ht]
\centering
\small
\begin{tabular}{l|cc|cc|cc}
\cline{1-7}
\rule{0pt}{0pt}\multirow{2}{*}{Application} & \multicolumn{2}{c|}{\textbf{\Large Rice31}} & \multicolumn{2}{c|}{\textbf{\Large Caltech36}} & \multicolumn{2}{c}{\textbf{\Large Cora}}\\ \cline{2-7} 
                             & 5\%     & 10\%  & 10\%     & 20\%  & 5\%     & 10\%   \\ \cline{1-7}\cline{1-7}
LP          & 80.80  & 87.51  & 70.38  & 78.03  & 69.68   & 74.49   \\
NHOLS       & 86.95  & 89.93  & 82.41  & 85.63  & 65.99   & 72.14   \\
\cline{1-7}
PL          & 58.19 $\pm$ 0.10  & 71.83 $\pm$ 0.32  & 41.08 $\pm$ 0.63  & 49.80 $\pm$ 0.51  & 46.94 $\pm$ 0.06   & 54.94 $\pm$ 0.04  \\ 
PL+C\&S     & 82.31 $\pm$ 0.10  & 88.51 $\pm$ 0.10  & 76.58 $\pm$ 1.56  & 79.83 $\pm$ 0.20  & 60.11 $\pm$ 0.11   & 79.02 $\pm$ 0.20  \\
PL+NLCS     & \textbf{87.44 $\pm$ 0.05}  & \textbf{90.77 $\pm$ 0.06}  & \textbf{81.88 $\pm$ 0.47}  & \textbf{84.62 $\pm$ 0.15}  & \textbf{77.22 $\pm$ 0.08}   & \textbf{80.71 $\pm$ 0.05}  \\
\cline{1-7}
MLP         & 69.33 $\pm$ 0.66  & 79.84 $\pm$ 0.61 & 46.93 $\pm$ 1.88  & 57.27 $\pm$ 1.81  & 58.74 $\pm$ 0.86   & 63.80 $\pm$ 0.81   \\
MLP+C\&S         & 72.72 $\pm$ 0.97  & 81.69 $\pm$ 0.67 & 72.49 $\pm$ 2.14  & 75.58 $\pm$ 1.23  & 75.32 $\pm$ 1.08   & \textbf{80.48 $\pm$ 0.74}   \\
MLP+NLCS         & \textbf{78.35 $\pm$ 0.90}  & \textbf{85.16 $\pm$ 0.46} & \textbf{80.58 $\pm$ 0.82}  & \textbf{83.92 $\pm$ 0.73}  & \textbf{75.76 $\pm$ 0.51}   & 80.26 $\pm$ 0.37   \\
\cline{1-7}
GAT         & 73.20 $\pm$ 0.27  & 81.15 $\pm$ 0.62  & 22.71 $\pm$ 1.30  & 36.00 $\pm$ 1.59  & 78.40 $\pm$ 0.32   & 80.93 $\pm$ 0.66   \\
GAT+C\&S         & 78.32 $\pm$ 0.70  & 84.86 $\pm$ 0.60  & 73.24 $\pm$ 2.96 & 74.82 $\pm$ 1.49  & 79.35 $\pm$ 0.46   & \textbf{81.51 $\pm$ 0.60}   \\
GAT+NLCS         & \textbf{82.65 $\pm$ 0.45}  & \textbf{87.09 $\pm$ 0.30}  & \textbf{82.96 $\pm$ 0.51}  & \textbf{80.99 $\pm$ 0.58}  & \textbf{79.51 $\pm$ 0.33}   & 81.26 $\pm$ 0.44   \\
\cline{1-7}
\end{tabular}
\begin{tabular}{l|cc|cc|cc}
\cline{1-7}
\rule{0pt}{0pt}\multirow{2}{*}{Application} & \multicolumn{2}{c|}{\textbf{\Large CiteSeer}} & \multicolumn{2}{c}{\textbf{\Large Arxiv}} & \multicolumn{2}{c}{\textbf{\Large PubMed}}\\ \cline{2-7} 
                             & 5\%     & 10\%     & 5\%     & 10\%   & 5\%     & 10\%  \\ \cline{1-7}\cline{1-7}
LP          & 49.01   & 53.43    & 65.48   & 67.62  & 79.12   & 79.94  \\
NHOLS       & 46.69   & 51.87    & 64.58   & 66.67  & 76.30   & 79.92  \\
\cline{1-7}
PL          & 52.28 $\pm$ 0.08   & 58.50 $\pm$ 0.19   & 48.53 $\pm$ 0.06   & 50.15 $\pm$ 0.02   & 76.91 $\pm$ 0.03   & 79.52 $\pm$ 0.03   \\ 
PL+C\&S     & 61.70 $\pm$ 0.11   & 64.80 $\pm$ 0.12   & \textbf{67.24 $\pm$ 0.04}   & \textbf{68.55 $\pm$ 0.01}   & \textbf{81.75 $\pm$ 0.04}   & \textbf{81.83 $\pm$ 0.01}   \\
PL+NLCS     & \textbf{62.81 $\pm$ 0.09}  & \textbf{65.64 $\pm$ 0.13}  & 65.72 $\pm$ 0.05   & 67.99 $\pm$ 0.01   & 80.76 $\pm$ 0.07   & 80.32 $\pm$ 0.06   \\
\cline{1-7}
MLP         & 50.00 $\pm$ 1.38   & 63.28 $\pm$ 0.83   & 68.33 $\pm$ 0.10   & 70.16 $\pm$ 0.05   & 79.16 $\pm$ 0.25   & 81.88 $\pm$ 0.33   \\
MLP+C\&S         & 59.72 $\pm$ 1.14   & \textbf{69.20 $\pm$ 0.94}     & \textbf{70.80 $\pm$ 0.08}   & \textbf{72.35 $\pm$ 0.07}   & \textbf{83.43 $\pm$ 0.25}   & \textbf{85.22 $\pm$ 0.22}   \\
MLP+NLCS         & \textbf{62.62 $\pm$ 0.57}   & 68.08 $\pm$ 0.89     & 69.39 $\pm$ 0.08   & 71.29 $\pm$ 0.13   & 82.67 $\pm$ 0.32   & 84.26 $\pm$ 0.15   \\
\cline{1-7}
GAT         & 64.26 $\pm$ 0.77   & 67.91 $\pm$ 0.49   & 70.13 $\pm$ 0.17   & 71.87 $\pm$ 0.08   & 84.20 $\pm$ 0.26   & 85.60 $\pm$ 0.25   \\
GAT+C\&S         & \textbf{65.34 $\pm$ 0.63}   & \textbf{67.48 $\pm$ 0.57}   & \textbf{71.24 $\pm$ 0.13}   & \textbf{72.69 $\pm$ 0.05}   & 84.34 $\pm$ 0.21   & 85.13 $\pm$ 0.12   \\
GAT+NLCS         & 64.94 $\pm$ 0.53   & 66.97 $\pm$ 0.58   & 70.44 $\pm$ 0.12   & 72.17 $\pm$ 0.04   & \textbf{84.43 $\pm$ 0.24}   & \textbf{85.62 $\pm$ 0.24}   \\
\cline{1-7}
\end{tabular}
\captionof{table}{The prediction accuracy on six datasets listed in Appendix \ref{exp-setup-app} with 5\% and 10\% of known labels. }
\label{tab:summary}
\end{table*}

\section{Experiment Evaluation}
\label{nlcs-exp}

In this section, we conduct empirical evaluations to assess the performance of the NLCS model. Our analysis aims to investigate the following questions: 

\textbf{Question 1:} Can the utilization of nonlinear higher-order graph representations improve performance compared to using only graph representations?

\textbf{Question 2:} How does the higher-order graph representation affect the prediction results in each step?

\textbf{Question 3:} How do the base prediction and post-processing steps influence each other, and what is the nature of their interaction?
\label{expset}

In order to answer these questions, NLCS is compared against C\&S over three base prediction models and six datasets. We also perform an in-depth analysis of each step in the NLCS model by analyzing the distribution of output values from both network models and post-processing steps. Additionally, we investigate the impact of post-process steps on base prediction via the training process. By conducting thorough experiments, we aim to gain insights into the performance and interpretability of the NLCS model.

\begin{table*}[ht]
\centering
\small
\begin{tabular}{l|cc|cc|cc}
\cline{1-7}
\rule{0pt}{0pt}\multirow{2}{*}{Application} & \multicolumn{2}{c|}{\textbf{\Large Rice31}} & \multicolumn{2}{c|}{\textbf{\Large Caltech36}} & \multicolumn{2}{c}{\textbf{\Large Cora}} \\ \cline{2-7} 
                             & 5\%     & 10\%     & 10\%     & 20\%     & 5\%     & 10\%      \\ \cline{1-7}\cline{1-7}
PL          & 58.19 $\pm$ 0.10  & 71.83 $\pm$ 0.32  & 41.08 $\pm$ 0.63  & 49.80 $\pm$ 0.51  & 46.94 $\pm$ 0.06   & 54.94 $\pm$ 0.04  \\ 
+linear correction     & 73.36 $\pm$ 0.10  & 84.34 $\pm$ 0.24  & 72.64 $\pm$ 2.28  & 77.41 $\pm$ 1.50  & 55.06 $\pm$ 0.11   & 68.80 $\pm$ 0.49  \\
+C\&S     & 82.31 $\pm$ 0.10  & 88.51 $\pm$ 0.10  & 76.58 $\pm$ 1.56  & 79.83 $\pm$ 0.20  & 60.11 $\pm$ 0.11   & 79.02 $\pm$ 0.20  \\
+nonlinear correction     & 80.57 $\pm$ 0.27  & 88.04 $\pm$ 0.24  & 71.47 $\pm$ 1.73  & 77.71 $\pm$ 1.78  & 54.96 $\pm$ 0.14   & 68.43 $\pm$ 0.55  \\
+NLCS     & \textbf{87.44 $\pm$ 0.05}  & \textbf{90.77 $\pm$ 0.06}  & \textbf{78.08 $\pm$ 0.78}  & \textbf{81.88 $\pm$ 0.47}  & \textbf{77.22 $\pm$ 0.08}   & \textbf{80.71 $\pm$ 0.05}  \\
\cline{1-7}
\end{tabular}
\begin{tabular}{l|cc|cc|cc}
\cline{1-7} 
\rule{0pt}{0pt}\multirow{2}{*}{Application} & \multicolumn{2}{c|}{\textbf{\Large CiteSeer}} & \multicolumn{2}{c}{\textbf{\Large Arxiv}} & \multicolumn{2}{c}{\textbf{\Large PubMed}}\\ \cline{2-7} 
                             & 5\%     & 10\%     & 5\%     & 10\%      & 5\%     & 10\%     \\ \cline{1-7} \cline{1-7} 
PL          & 52.28 $\pm$ 0.08   & 58.50 $\pm$ 0.19    & 48.53 $\pm$ 0.06   & 50.15 $\pm$ 0.02   & 76.91 $\pm$ 0.03   & 79.52 $\pm$ 0.03   \\ 
+linear correction     & 58.62 $\pm$ 0.08  & 62.76 $\pm$ 0.04  & 61.43 $\pm$ 0.20   & 63.88 $\pm$ 0.02  & 81.88 $\pm$ 0.09   & 82.01 $\pm$ 0.04   \\
+C\&S     & 61.70 $\pm$ 0.11   & 64.80 $\pm$ 0.12   & \textbf{67.24 $\pm$ 0.04}   & \textbf{68.55 $\pm$ 0.01}   & \textbf{81.75 $\pm$ 0.04}   & \textbf{81.83 $\pm$ 0.01}   \\
+nonlinear correction     & 59.83 $\pm$ 0.12  & 63.91 $\pm$ 0.08  & 57.27 $\pm$ 0.19   & 61.09 $\pm$ 0.02   & 78.90 $\pm$ 0.08   & 79.21 $\pm$ 0.05   \\
+NLCS     & \textbf{62.81 $\pm$ 0.09}  & \textbf{65.64 $\pm$ 0.13}  & 65.72 $\pm$ 0.05   & 67.99 $\pm$ 0.01   & 80.76 $\pm$ 0.07   & 80.32 $\pm$ 0.06   \\
\cline{1-7} 
\end{tabular}
\captionof{table}{The prediction accuracy on each step of the post-processing is compared between linear and nonlinear.}
\label{tab:cs-summary}
\end{table*}

\subsection{Experiment setup}
\label{exp-setup}
We compare NLCS to several baselines including propagation algorithms and network models: standard label propagation (LP) \cite{zhou2003learning}, nonlinear higher-order label spreading (NHOLS) \cite{tudisco2021nonlinear}, Plain linear (PL), Muti-layer perceptron (MLP), and Graph attention network (GAT) \cite{velivckovic2017graph}. For each baseline method, we apply both our post-processing step and C\&S to analyze their performance via six commonly used datasets. More details regarding baseline models, datasets, and parameters are included in Appendix \ref{exp-setup-app}. We provide the source code link\footnote{Code Repository: \url{https://drive.google.com/file/d/1HCdZ63O6RBQbdjoKfFNDTdyx9YTrAVzt/view?usp=drive_link}} for our experiments.

\begin{figure}[h]
    \centering
    \includegraphics[width=0.98\linewidth]{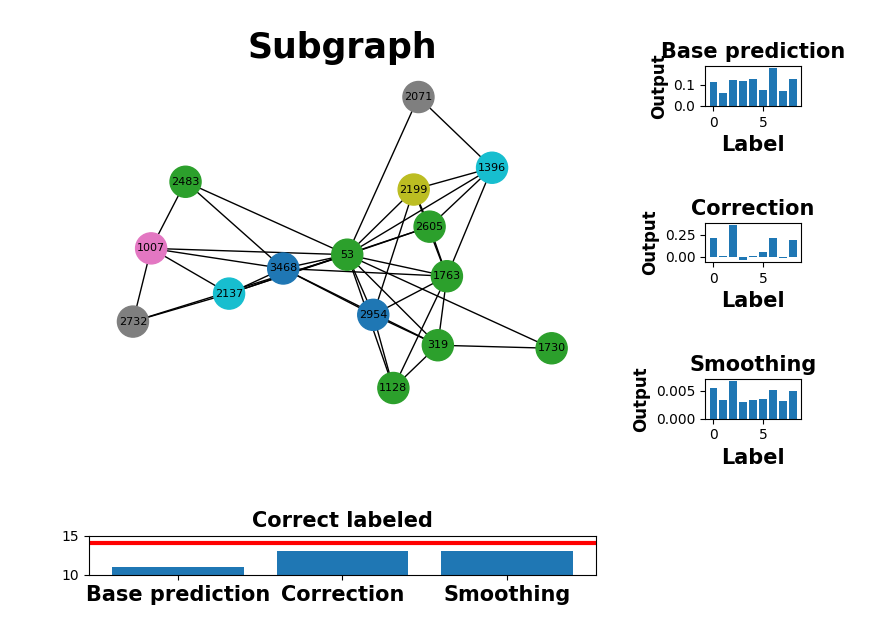}
  \caption{The subgraph visualization showcases a center node 53 and its one-hop neighbors from the Rice31 dataset with 10\% known labels using PL base prediction.}
  \label{fig-rice} 
\end{figure}

\begin{figure}[h]
    \centering
    \subfloat[\label{fig-rice-bins-0.5}Rice31-5\%]{
    \includegraphics[width=0.48\linewidth]{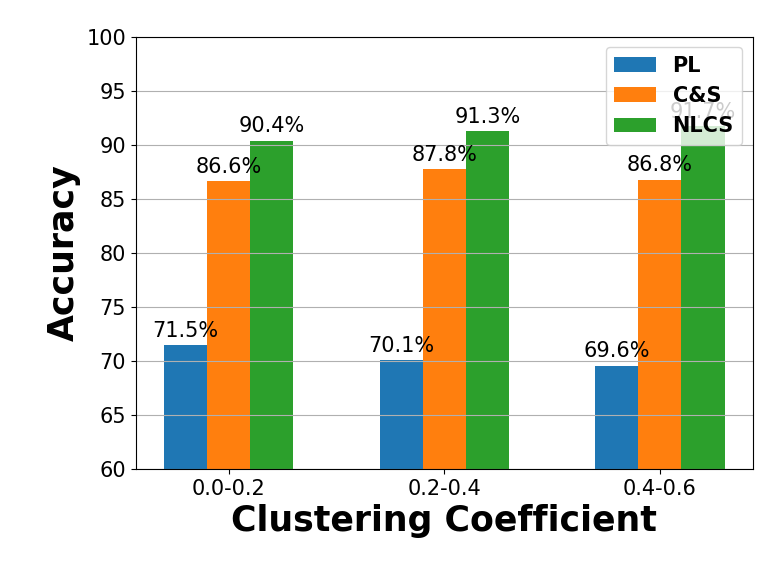}}
    \hfill
    \subfloat[\label{fig-rice-bins-0.1}Rice31-10\%]{
    \includegraphics[width=0.48\linewidth]{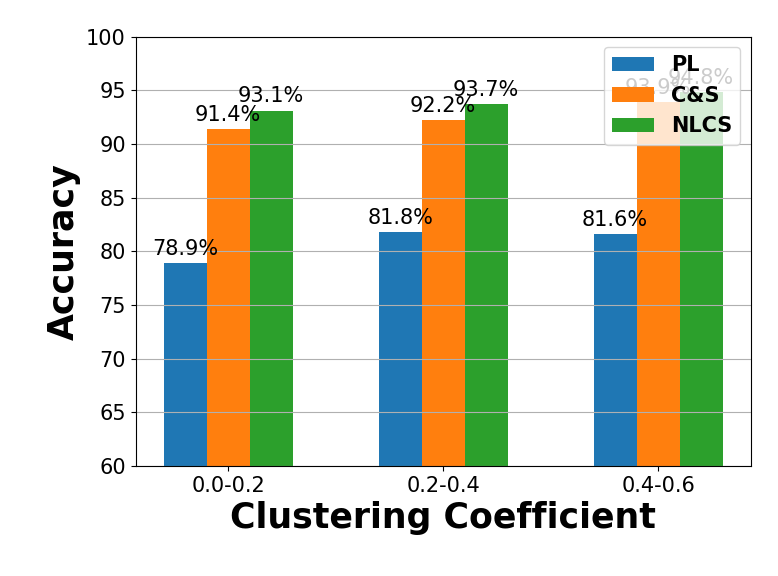}}
    \hfill
    \subfloat[\label{fig-cora-bins-0.5}Cora-5\%]{
    \includegraphics[width=0.48\linewidth]{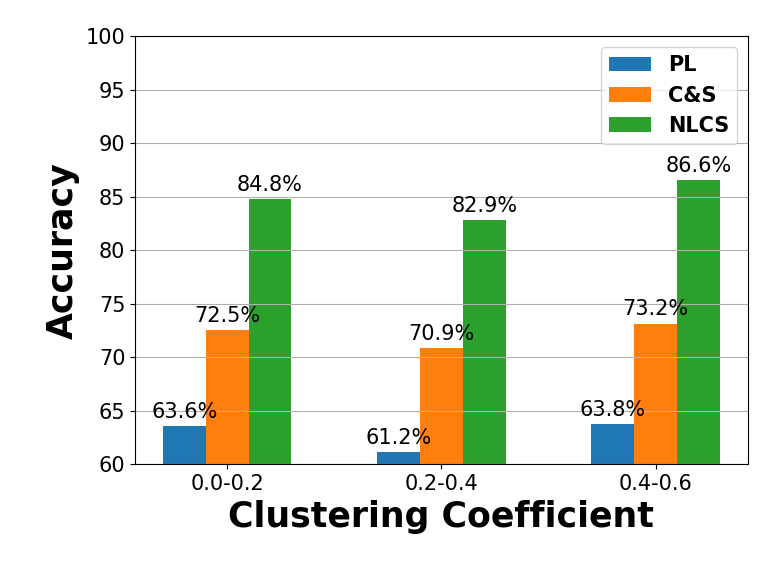}}
    \hfill
    \subfloat[\label{fig-cora-bins-0.1}Cora-10\%]{
    \includegraphics[width=0.48\linewidth]{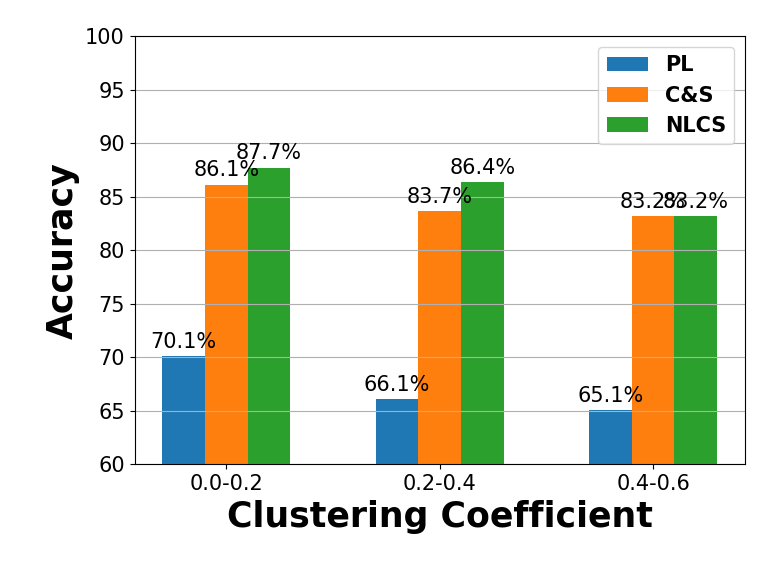}}

  \caption{The prediction accuracy at different ranges of clustering coefficient. }
  \label{fig-bins} 
\end{figure}

\begin{figure}[h]
    \centering
    \subfloat[\label{fig-rice-base} Rice31]{
    \includegraphics[width=0.48\linewidth]{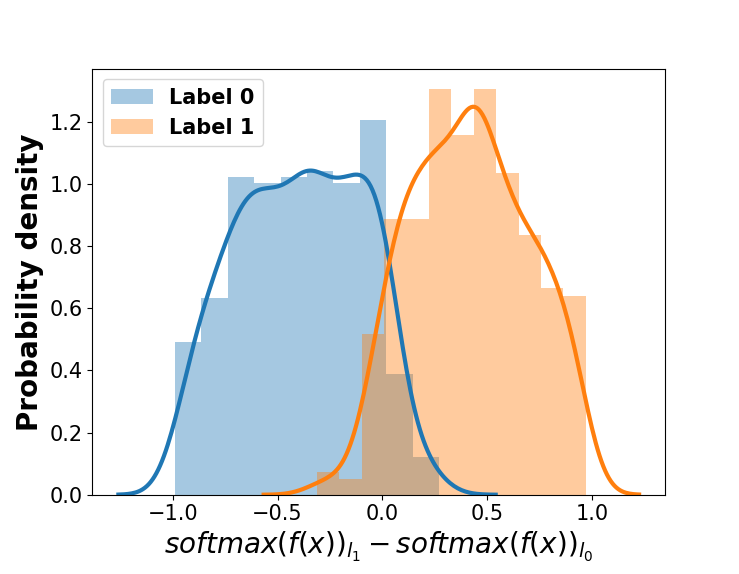}}
    \hfill
    \subfloat[\label{fig-rice-corr} Rice31]{
    \includegraphics[width=0.48\linewidth]{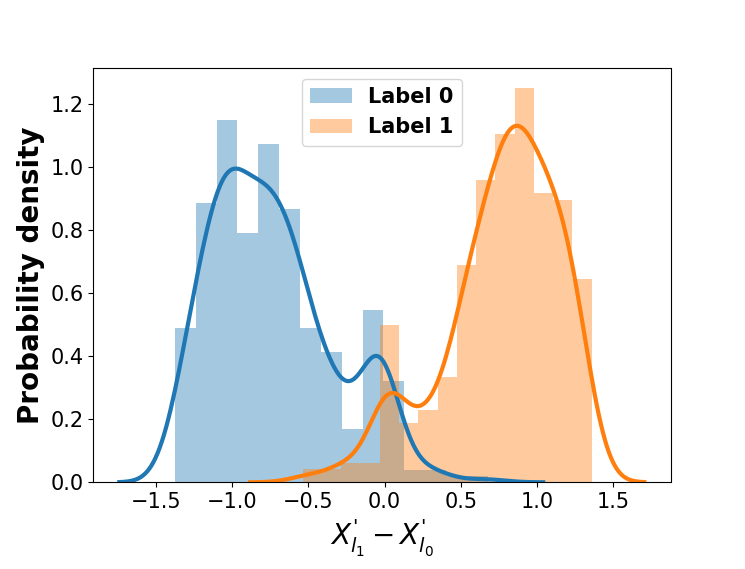}}
    \hfill
    \subfloat[\label{fig-cora-base} Cora]{
    \includegraphics[width=0.48\linewidth]{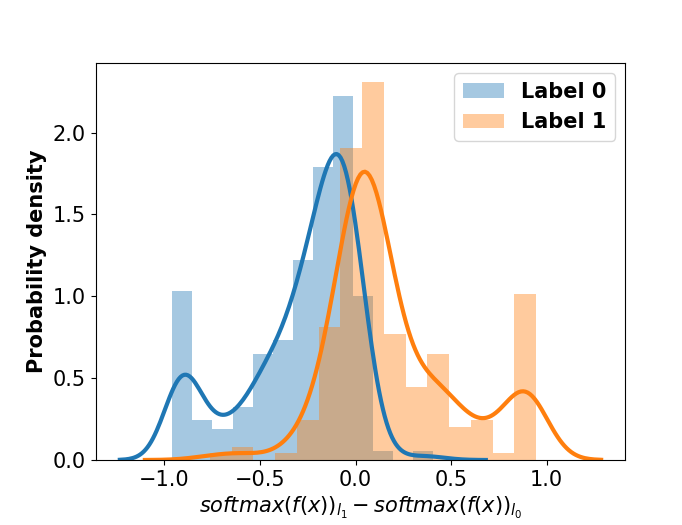}}
    \hfill
    \subfloat[\label{fig-cora-corr} Cora]{
    \includegraphics[width=0.48\linewidth]{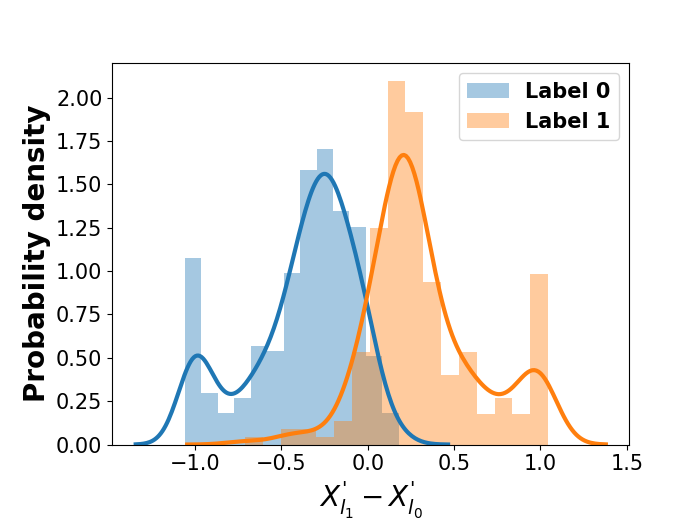}}
    \hfill
   \caption{Demonstrate the distribution of the difference value between the first two output values regarding base prediction and correction for the Rice31 and Cora datasets. }
   \label{fig-cs} 
\end{figure}

\begin{figure}[h]
    \centering
    \begin{minipage}{.48\linewidth}
        \centering
        \subfloat[\label{fig-fout-pca}]{
        \includegraphics[width=1\linewidth]{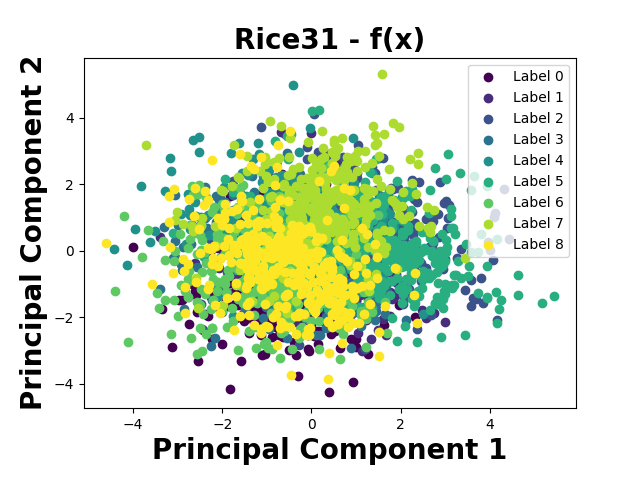}}
    \end{minipage}
    \begin{minipage}{.48\linewidth}
        \centering
        \subfloat[\label{fig-model-pca}]{
        \includegraphics[width=1\linewidth]{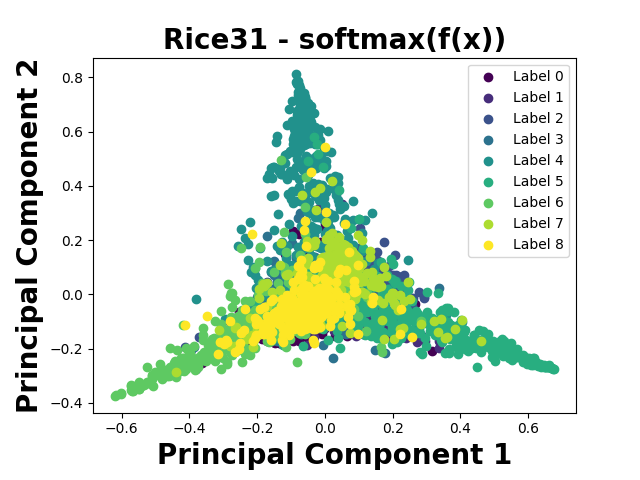}}
    \end{minipage}\par\medskip
    \begin{minipage}{.48\linewidth}
        \centering
        \subfloat[\label{fig-correction-pca}]{
        \includegraphics[width=1\linewidth]{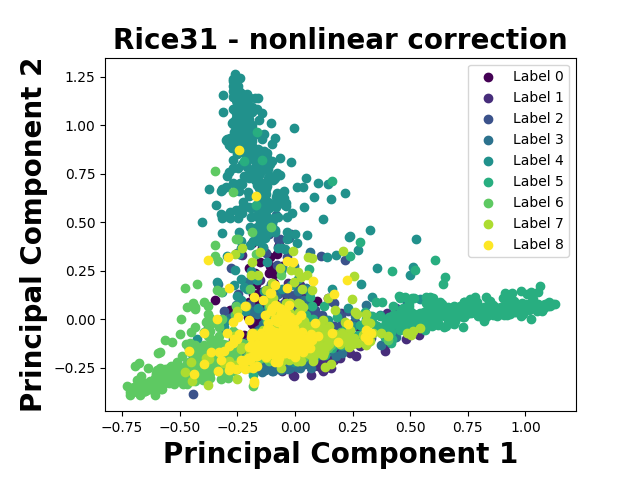}}
    \end{minipage}
    \begin{minipage}{.48\linewidth}
        \centering
        \subfloat[\label{fig-smooth-pca}]{
        \includegraphics[width=1\linewidth]{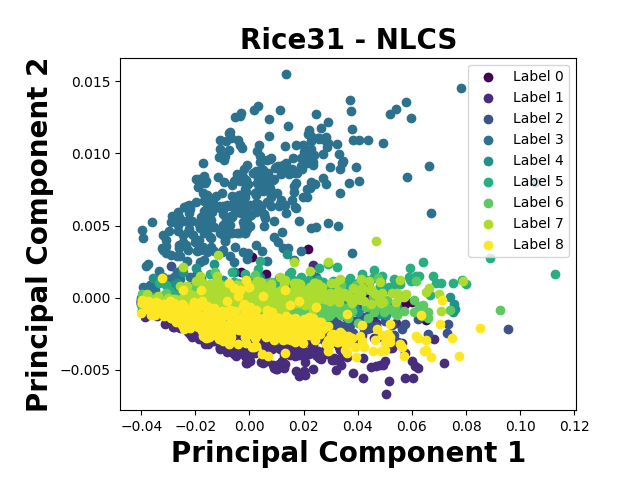}}
    \end{minipage}
    \caption{The PCA visualization on the output value of Rice31 with 5\% known labels. }
    \label{fig-pca} 
\end{figure}

\subsection{Performance evaluation}
We compare the performance of base predictions after applying our post-processing step with C\&S and other neural network models. As shown in Table \ref{tab:summary}, NLCS improves the average accuracy by 19.77\%, 11.14\%, and 10.22\% across all datasets when compared to PL, MLP, and GAT base predictions, respectively. On the other hand, the C\&S improves the average accuracy by 16.96\%, 9.19\%, and 8.50\% across all datasets with PL, MLP, and GAT base prediction. The plain linear network model exhibits low accuracy on all datasets when the node features are unavailable as expected. Meanwhile, the MLP and GAT effectively learn the node feature representations for classification. 
Notably, the performance of two social network datasets is further improved by 5.46\% when applying our nonlinear higher-order error spreading compared to the C\&S across all three base predictions. This improvement is particularly significant because social network datasets lack node feature information, limiting the performance of MLP and GAT models. In this case, the combination of base prediction with nonlinear post-processing surpasses the performance of sophisticated neural network models.
In the case of citation networks, the NLCS achieves an average improvement of 5.19\% on Cora and CiteSeer compared to C\&S with PL. The performance of NLCS is competitive against C\&S using MLP and GAT base predictions. However, the NLCS is slightly lower than C\&S with PL base prediction by 1.15\% in Arxiv and PubMed datasets. Overall, the NLCS model improves the performance by 8.1\% across all three base prediction models. The post-processing step yields greater improvements when the node features are not directly associated with labels, and the network models struggle to learn the representation of nodes. This observation aligns with the notation of feature and label influence that has been discussed in \cite{wang2021combining}. Overall, NLCS shows its effectiveness by yielding lower standard deviation (STD) in 72.2\% of cases and higher accuracy in 58.3\% of cases compared to C\&S. This indicates that NLCS provides more consistent and reliable results with improved prediction accuracy utilizing higher-order representation.

Furthermore, we investigate the impact of the nonlinear function in each step of the post-processing. As shown in Table \ref{tab:cs-summary}, it demonstrates the accuracy of PL base prediction, after correction, and after smoothing compared to the linear-based C\&S. The nonlinear correction improves the accuracy by 12.73\% compared to the improvement of 12.79\% achieved by linear correction. Moreover, nonlinear smoothing improves the accuracy by 6.50\%, while linear smoothing only improves it by 4.17\%. In summary, the nonlinear function leads to better performance in the final prediction.

\subsubsection{Distribution Analysis}
To gain insights into the effectiveness of the post-processing step, we conducted an output value distribution analysis at individual nodes and their neighbors. As shown in Figure \ref{fig-rice}, we select an individual node (Node 53) from the Rice31 dataset and illustrate the output value distribution of the center node regarding different classes at each step on the right-hand side (base prediction, nonlinear correction, and nonlinear smoothing, respectively). Additionally, we provide the number of correct predictions for its adjacency nodes on the bottom, where the red line indicates the total number of neighbors. Based on the observation of this subgraph, it is obvious that the neighbors of this individual node belong to multiple classes, which brings a challenge for accurate predictions. Remarkably, for this individual node, even though both the node itself and its neighbors are unlabeled, our method successfully predicts 13 out of 14 neighbors in this subgraph. This success is due to the prevalent triangular structure observed in the subgraph, which our method utilizes effectively by leveraging higher-order representation to improve prediction accuracy. Moreover, we analyze the prediction accuracy corresponding to the ranges of clustering coefficient from 0 to 0.6 in Figure \ref{fig-bins}. This range corresponds to the majority of nodes in the dataset. Notably, NLCS consistently outperformed both the base prediction and C\&S methods across different coefficients. Overall, by incorporating higher-order representation in the correction and smoothing framework, our method shows modeling flexibility in complex relationships while improving performance constantly.

\begin{figure}[h]
    \centering
    \subfloat[\label{fig-rice-trend}]{
    \includegraphics[width=0.48\linewidth]{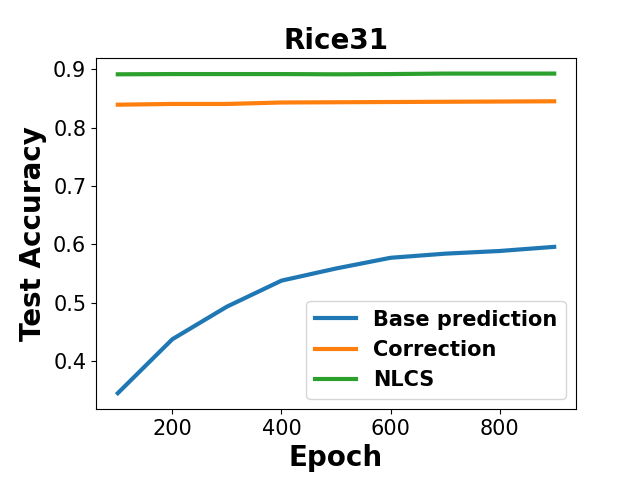}}
    \hfill
    \subfloat[\label{fig-cora-trend}]{
    \includegraphics[width=0.48\linewidth]{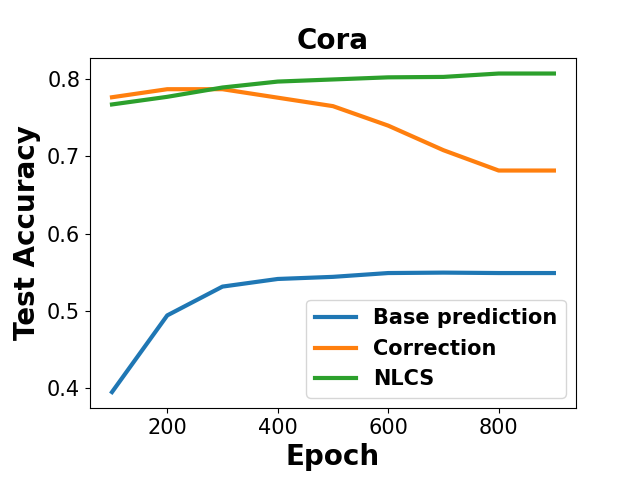}}
    \hfill
    \subfloat[\label{fig-citeseer-trend}]{
    \includegraphics[width=0.48\linewidth]{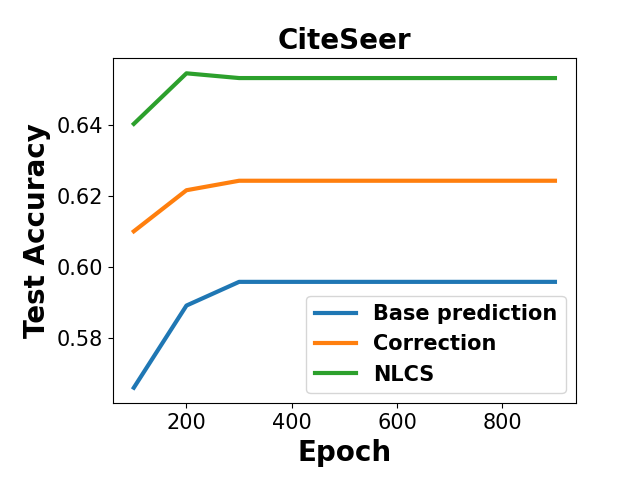}}
    \hfill
    \subfloat[\label{fig-pubmed-trend}]{
    \includegraphics[width=0.48\linewidth]{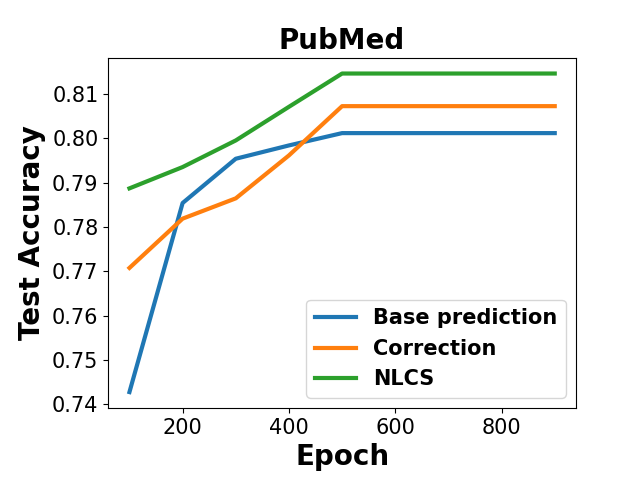}}
    \hfill
  \caption{The accuracy during the training to the base prediction model, nonlinear correction, and NLCS. }
  \label{fig-trend} 
\end{figure}

We further analyzed the output value distribution from an overall perspective at each stage of our method: base prediction, correction, and smoothing. In Figure \ref{fig-cs}, we observe the classification change by selecting the first two classes of nodes and visualizing the difference between the first two output values. Figure \ref{fig-rice-base} and \ref{fig-cora-base} illustrate that the base prediction is capable of distinguishing between classes based on the feature vectors. However, the misclassification happens when the difference value is greater than 0 in Label 0 and smaller than 0 in Label 1. Figure \ref{fig-rice-corr} - \ref{fig-cora-corr} (and Figure \ref{fig-cs-app} in appendix \ref{add-exp}) demonstrates that the overlapping area decreases after correction and smoothing, indicating a reduction in incorrect classifications and an improvement in accuracy. This trend is more significant in the analysis of the Cora dataset, as shown in Figure \ref{fig-cora-base} to \ref{fig-cora-cs}.

To intuitively understand how our method clusters the nodes in the spatial domain, we visualize the multidimensional output vectors of prediction using principal component analysis (PCA). The results of Figure \ref{fig-pca} demonstrate the 2-dimensional plots of the output value from linear layer, PL model, correction, and smoothing. The 3-dimensional plots using 3 principal components are in Appendix \ref{add-exp} Figure \ref{fig-pca-app}. The results show that the neural network learns the representation of nodes and provides a baseline of accuracy. Subsequently, the nonlinear correction further brings the cluster to a more confident level, and the nonlinear smoothing compresses the information into a lower dimension.

\subsubsection{Accuracy timeline}

We analyzed the accuracy during the training process of the base prediction and after applying the NLCS post-processing step to assess performance stability. Figure \ref{fig-trend} shows the testing accuracy during the training and applies the NLCS on every 100 epochs. The plots show that the post-processing step consistently improves the performance over time. Although the specific benefits gained from the correction and smoothing steps may vary depending on the dataset, the overall performance remains unaffected. This suggests that the NLCS is robust and stable, even when the base prediction is not well-trained. 

\section{Conclusion}
We have combined nonlinearity and higher-order representation into the existing post-processing framework to utilize labels and features jointly. Delving into this unexplored domain of joint utilization illustrates the effectiveness of higher-order information, which also offers insights into how different methods leverage the relationships in graphs. We notice that nonlinear and linear methods both exhibit varying degrees of effectiveness across datasets. This observation suggests developing a unified framework that can recognize the property of the local graph more effectively and provide adaptive solutions for distinct datasets, which is being explored.

\clearpage

\bibliographystyle{unsrt}
\bibliography{reference}

\begin{thebibliography}{10}

\bibitem{butts2008social}
Carter~T Butts.
\newblock Social network analysis: A methodological introduction.
\newblock {\em Asian Journal of Social Psychology}, 11(1):13--41, 2008.

\bibitem{dawson2014current}
Shane Dawson, Dragan Ga{\v{s}}evi{\'c}, George Siemens, and Srecko Joksimovic.
\newblock Current state and future trends: A citation network analysis of the
  learning analytics field.
\newblock In {\em Proceedings of the fourth international conference on
  learning analytics and knowledge}, pages 231--240, 2014.

\bibitem{adeniyi2016automated}
David~Adedayo Adeniyi, Zhaoqiang Wei, and Yang Yongquan.
\newblock Automated web usage data mining and recommendation system using
  k-nearest neighbor (knn) classification method.
\newblock {\em Applied Computing and Informatics}, 12(1):90--108, 2016.

\bibitem{zhu2020beyond}
Jiong Zhu, Yujun Yan, Lingxiao Zhao, Mark Heimann, Leman Akoglu, and Danai
  Koutra.
\newblock Beyond homophily in graph neural networks: Current limitations and
  effective designs.
\newblock {\em Advances in neural information processing systems},
  33:7793--7804, 2020.

\bibitem{zhou2003learning}
Dengyong Zhou, Olivier Bousquet, Thomas Lal, Jason Weston, and Bernhard
  Sch{\"o}lkopf.
\newblock Learning with local and global consistency.
\newblock {\em Advances in neural information processing systems}, 16, 2003.

\bibitem{huang2020combining}
Qian Huang, Horace He, Abhay Singh, Ser-Nam Lim, and Austin~R Benson.
\newblock Combining label propagation and simple models out-performs graph
  neural networks.
\newblock {\em arXiv preprint arXiv:2010.13993}, 2020.

\bibitem{verma2021graphmix}
Vikas Verma, Meng Qu, Kenji Kawaguchi, Alex Lamb, Yoshua Bengio, Juho Kannala,
  and Jian Tang.
\newblock Graphmix: Improved training of gnns for semi-supervised learning.
\newblock In {\em Proceedings of the AAAI conference on artificial
  intelligence}, volume~35, pages 10024--10032, 2021.

\bibitem{ding2022meta}
Kaize Ding, Jianling Wang, James Caverlee, and Huan Liu.
\newblock Meta propagation networks for graph few-shot semi-supervised
  learning.
\newblock In {\em Proceedings of the AAAI Conference on Artificial
  Intelligence}, volume~36, pages 6524--6531, 2022.

\bibitem{szummer2001partially}
Martin Szummer and Tommi Jaakkola.
\newblock Partially labeled classification with markov random walks.
\newblock {\em Advances in neural information processing systems}, 14, 2001.

\bibitem{tudisco2021nonlinear}
Francesco Tudisco, Austin~R Benson, and Konstantin Prokopchik.
\newblock Nonlinear higher-order label spreading.
\newblock In {\em Proceedings of the Web Conference 2021}, pages 2402--2413,
  2021.

\bibitem{scarselli2008graph}
Franco Scarselli, Marco Gori, Ah~Chung Tsoi, Markus Hagenbuchner, and Gabriele
  Monfardini.
\newblock The graph neural network model.
\newblock {\em IEEE transactions on neural networks}, 20(1):61--80, 2008.

\bibitem{li2018deeper}
Qimai Li, Zhichao Han, and Xiao-Ming Wu.
\newblock Deeper insights into graph convolutional networks for semi-supervised
  learning.
\newblock In {\em Thirty-Second AAAI conference on artificial intelligence},
  2018.

\bibitem{feng2020graph}
Wenzheng Feng, Jie Zhang, Yuxiao Dong, Yu~Han, Huanbo Luan, Qian Xu, Qiang
  Yang, Evgeny Kharlamov, and Jie Tang.
\newblock Graph random neural networks for semi-supervised learning on graphs.
\newblock {\em Advances in neural information processing systems},
  33:22092--22103, 2020.

\bibitem{oono2019graph}
Kenta Oono and Taiji Suzuki.
\newblock Graph neural networks exponentially lose expressive power for node
  classification.
\newblock {\em arXiv preprint arXiv:1905.10947}, 2019.

\bibitem{li2019deepgcns}
Guohao Li, Matthias Muller, Ali Thabet, and Bernard Ghanem.
\newblock Deepgcns: Can gcns go as deep as cnns?
\newblock In {\em Proceedings of the IEEE/CVF international conference on
  computer vision}, pages 9267--9276, 2019.

\bibitem{wang2021combining}
Hongwei Wang and Jure Leskovec.
\newblock Combining graph convolutional neural networks and label propagation.
\newblock {\em ACM Transactions on Information Systems (TOIS)}, 40(4):1--27,
  2021.

\bibitem{battaglia2018relational}
Peter~W Battaglia, Jessica~B Hamrick, Victor Bapst, Alvaro Sanchez-Gonzalez,
  Vinicius Zambaldi, Mateusz Malinowski, Andrea Tacchetti, David Raposo, Adam
  Santoro, Ryan Faulkner, et~al.
\newblock Relational inductive biases, deep learning, and graph networks.
\newblock {\em arXiv preprint arXiv:1806.01261}, 2018.

\bibitem{chong2020graph}
Yanwen Chong, Yun Ding, Qing Yan, and Shaoming Pan.
\newblock Graph-based semi-supervised learning: A review.
\newblock {\em Neurocomputing}, 408:216--230, 2020.

\bibitem{thekumparampil2018attention}
Kiran~K Thekumparampil, Chong Wang, Sewoong Oh, and Li-Jia Li.
\newblock Attention-based graph neural network for semi-supervised learning.
\newblock {\em arXiv preprint arXiv:1803.03735}, 2018.

\bibitem{chen2020simple}
Ming Chen, Zhewei Wei, Zengfeng Huang, Bolin Ding, and Yaliang Li.
\newblock Simple and deep graph convolutional networks.
\newblock In {\em International conference on machine learning}, pages
  1725--1735. PMLR, 2020.

\bibitem{zhou2021understanding}
Kuangqi Zhou, Yanfei Dong, Kaixin Wang, Wee~Sun Lee, Bryan Hooi, Huan Xu, and
  Jiashi Feng.
\newblock Understanding and resolving performance degradation in deep graph
  convolutional networks.
\newblock In {\em Proceedings of the 30th ACM International Conference on
  Information \& Knowledge Management}, pages 2728--2737, 2021.

\bibitem{huang2020tackling}
Wenbing Huang, Yu~Rong, Tingyang Xu, Fuchun Sun, and Junzhou Huang.
\newblock Tackling over-smoothing for general graph convolutional networks.
\newblock {\em arXiv preprint arXiv:2008.09864}, 2020.

\bibitem{chen2022bag}
Tianlong Chen, Kaixiong Zhou, Keyu Duan, Wenqing Zheng, Peihao Wang, Xia Hu,
  and Zhangyang Wang.
\newblock Bag of tricks for training deeper graph neural networks: A
  comprehensive benchmark study.
\newblock {\em IEEE Transactions on Pattern Analysis and Machine Intelligence},
  2022.

\bibitem{jia2022unifying}
Junteng Jia and Austin~R Benson.
\newblock A unifying generative model for graph learning algorithms: Label
  propagation, graph convolutions, and combinations.
\newblock {\em SIAM Journal on Mathematics of Data Science}, 4(1):100--125,
  2022.

\bibitem{eswaran2020higher}
Dhivya Eswaran, Srijan Kumar, and Christos Faloutsos.
\newblock Higher-order label homogeneity and spreading in graphs.
\newblock In {\em Proceedings of The Web Conference 2020}, pages 2493--2499,
  2020.

\bibitem{jia2020residual}
Junteng Jia and Austion~R Benson.
\newblock Residual correlation in graph neural network regression.
\newblock In {\em Proceedings of the 26th ACM SIGKDD International Conference
  on Knowledge Discovery \& Data Mining}, pages 588--598, 2020.

\bibitem{velivckovic2017graph}
Petar Veli{\v{c}}kovi{\'c}, Guillem Cucurull, Arantxa Casanova, Adriana Romero,
  Pietro Lio, and Yoshua Bengio.
\newblock Graph attention networks.
\newblock {\em arXiv preprint arXiv:1710.10903}, 2017.

\end{thebibliography}

\clearpage
\appendix
\section{Experiment setup}
\label{exp-setup-app}
\textbf{Baselines:} We compare NLCS to several baselines including propagation algorithms and neural network models: standard label propagation (LP) \cite{zhou2003learning}, nonlinear higher-order label spreading (NHOLS) \cite{tudisco2021nonlinear}, and following network models as base predictions:

\begin{itemize}
    \item Plain linear (PL): this network consists of one linear layer with softmax and uses spectral embedding as input. 
    \item Muti-layer perceptron (MLP): this network includes three linear layers with ReLu activation function with batch normalization, dropout, and softmax, which uses node feature vectors as input if available. 
    \item Graph attention network (GAT) \cite{velivckovic2017graph}: this network consists of three layers of convolution operation, which uses node features as input if available. 
\end{itemize}

For each base prediction, we train on 10 random initialized network models and apply both our post-processing step and C\&S to analyze their performance. 

\begin{table}[hb]
    \centering
    \begin{tabular}{l l l l l}
      \hline\hline 
      Application & Dataset & $|\mathcal{N}|$ & $|E|$ & $|L|$\\ [0.5ex] 
      \hline 
      Social & Rice31 & 3,560 & 317,828 & 9\\
      Social & Caltech36 & 590 & 25,644 & 8\\
      Citation & arxiv & 169,343 & 2,315,598 & 40\\
      Citation & Cora & 2,708 & 10,556 & 7\\
      Citation & CiteSeer & 3,327 & 9,104 & 6\\
      Citation & PubMed & 19,717 & 88,648 & 3\\
      \hline 
    \end{tabular} \\
    \caption{Dataset and graph size.}\label{tab:dataset}
\end{table}

\textbf{Datasets:} 
We evaluate the performance of each baseline using the datasets listed in Table \ref{tab:dataset}. For each dataset, we split it into train, validation, and test subsets based on the ratio of known labels ($k$). The split is performed by randomly selecting an equal number of samples from each class, with ratio $k$, $\frac{(1-k)}{2}$,  and $\frac{(1-k)}{2}$ respectively. The same split is applied consistently across different baselines for a fair comparison. Considering the size of Caltech36, the percentages of known labels are set to 10\% and 20\% in Table \ref{tab:summary}, while the other dataset remains 5\% and 10\%. It should be noted that the social network datasets do not include node features, whereas the citation network datasets are the opposite. 

\textbf{Parameters:}
We systematically explore the constant parameters $\alpha$ and $\beta$ in Eq. \ref{eq:nlc} and \ref{eq:lp} by varying them from range 0 to 1, with increments of 0.1. The provided source code includes the optimal parameter settings that we have discovered through extensive experimentation. Additionally, for the propagation function, we employ a fixed constant number of iterations ($t=50$), which has proven to be sufficient for convergence. Regarding the training of neural network models, all experiments run for 1000 epochs with learning rates ranging from 0.01 to 0.001, depending on the specific dataset. The multi-layer models (MLP and GAT) consist of 256 hidden neurons and a dropout rate of 0.5.

\section{Additional results}
\label{add-exp}
\begin{figure*}[h]
    \centering
    \subfloat[\label{fig-rice-10-NLCS} Rice31-NLCS]{
    \includegraphics[width=0.48\linewidth]{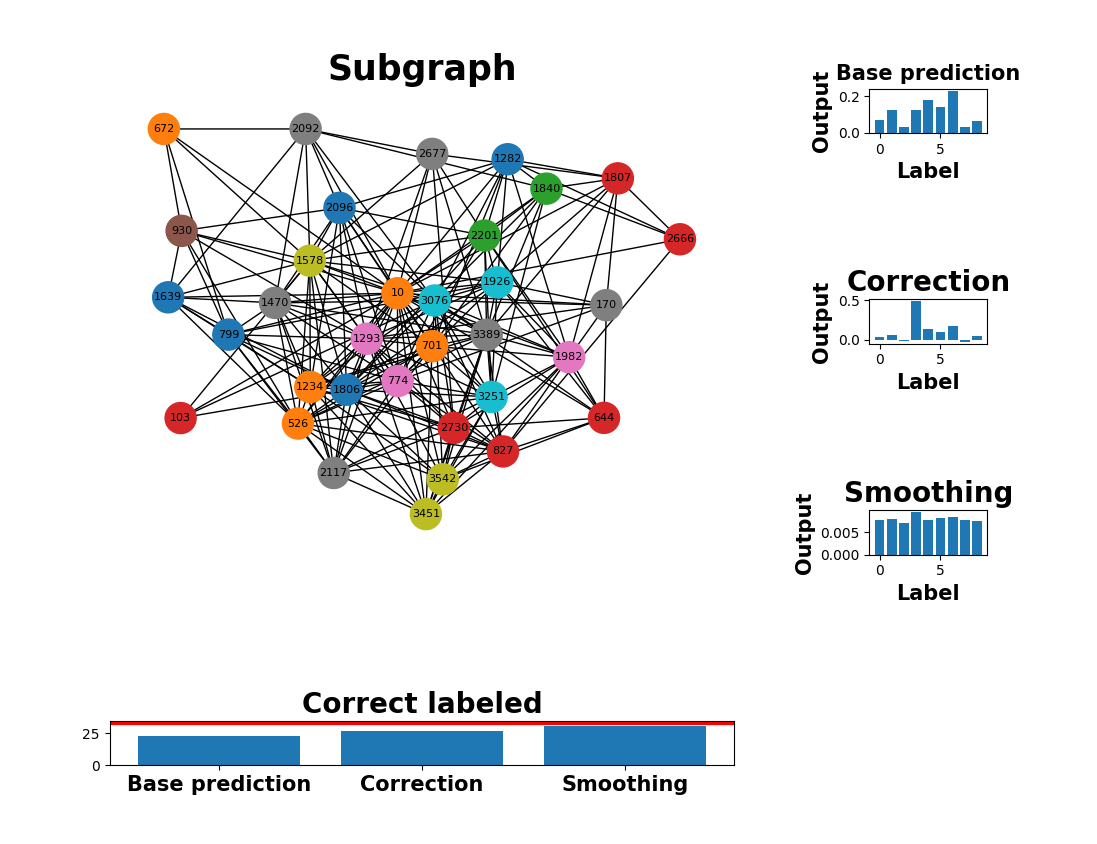}}
    \hfill
    \subfloat[\label{fig-rice-10-CS} Rice31-C\&S]{
    \includegraphics[width=0.48\linewidth]{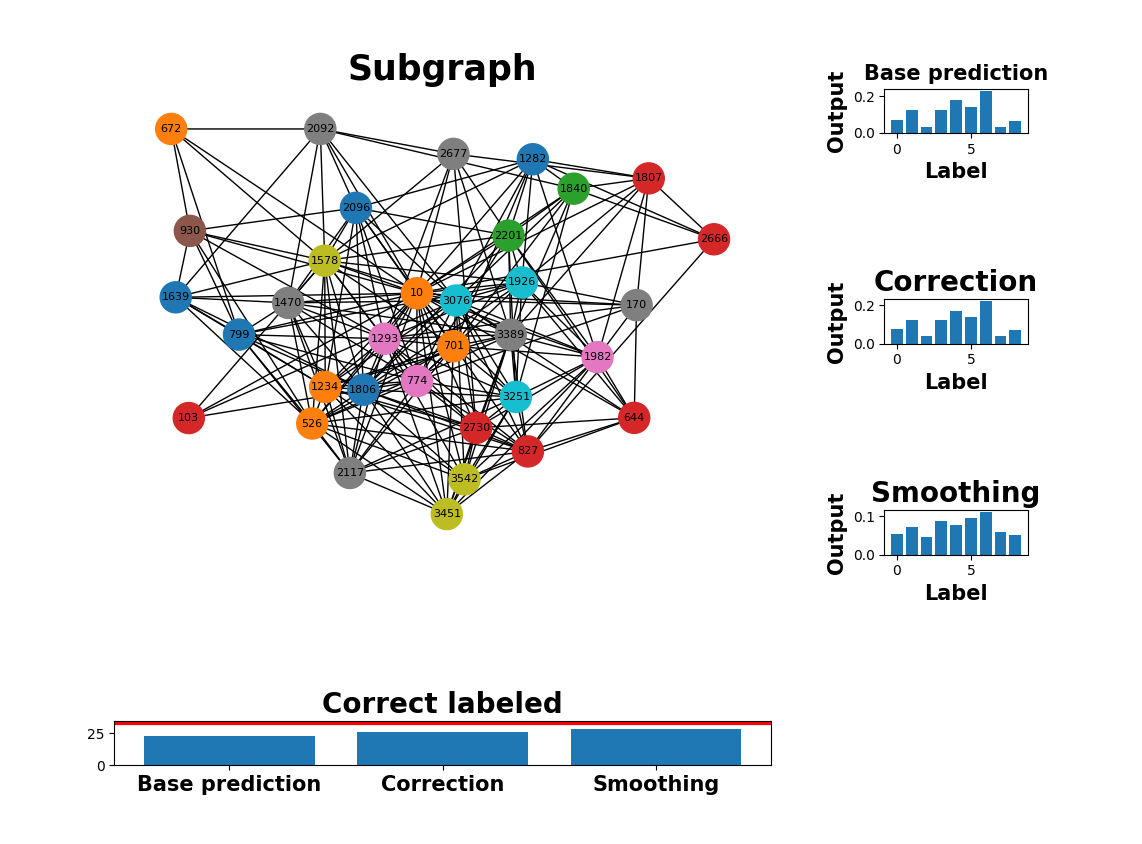}}
    \hfill
   \caption{The subgraph visualization showcases a center node 10 and its one-hop neighbors from Rice31 with 10\% known labels using PL base prediction.}
   \label{fig-rice-10-app} 
\end{figure*}

\begin{figure*}[h]
    \centering
    \subfloat[\label{fig-rice-171-NLCS} Rice31-NLCS]{
    \includegraphics[width=0.48\linewidth]{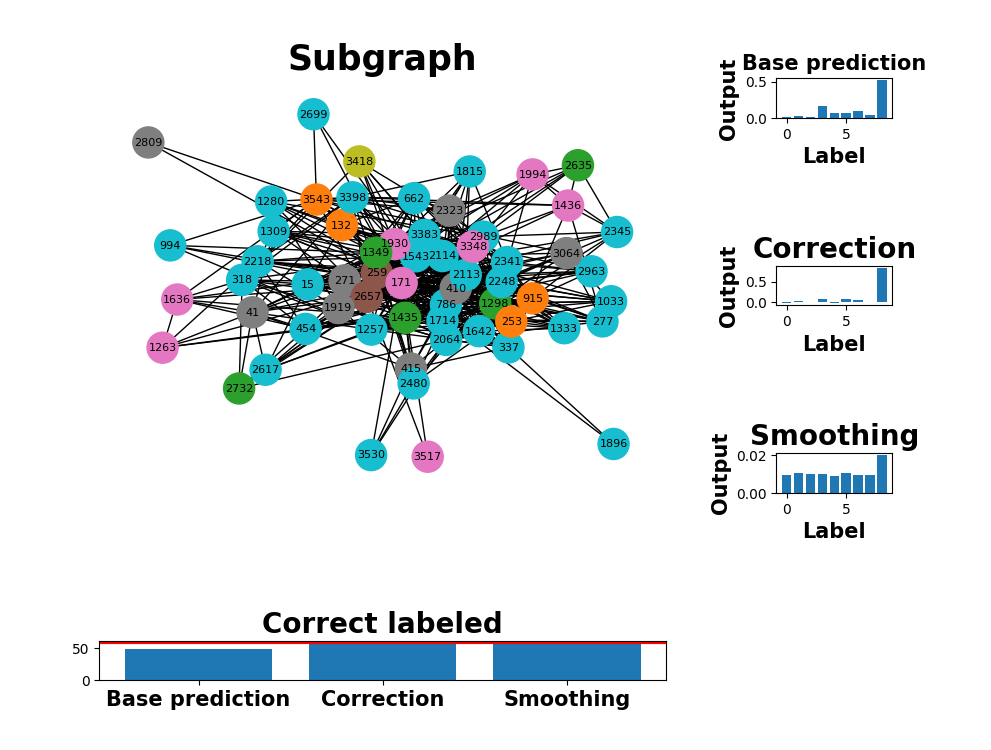}}
    \hfill
    \subfloat[\label{fig-rice-171-CS} Rice31-C\&S]{
    \includegraphics[width=0.48\linewidth]{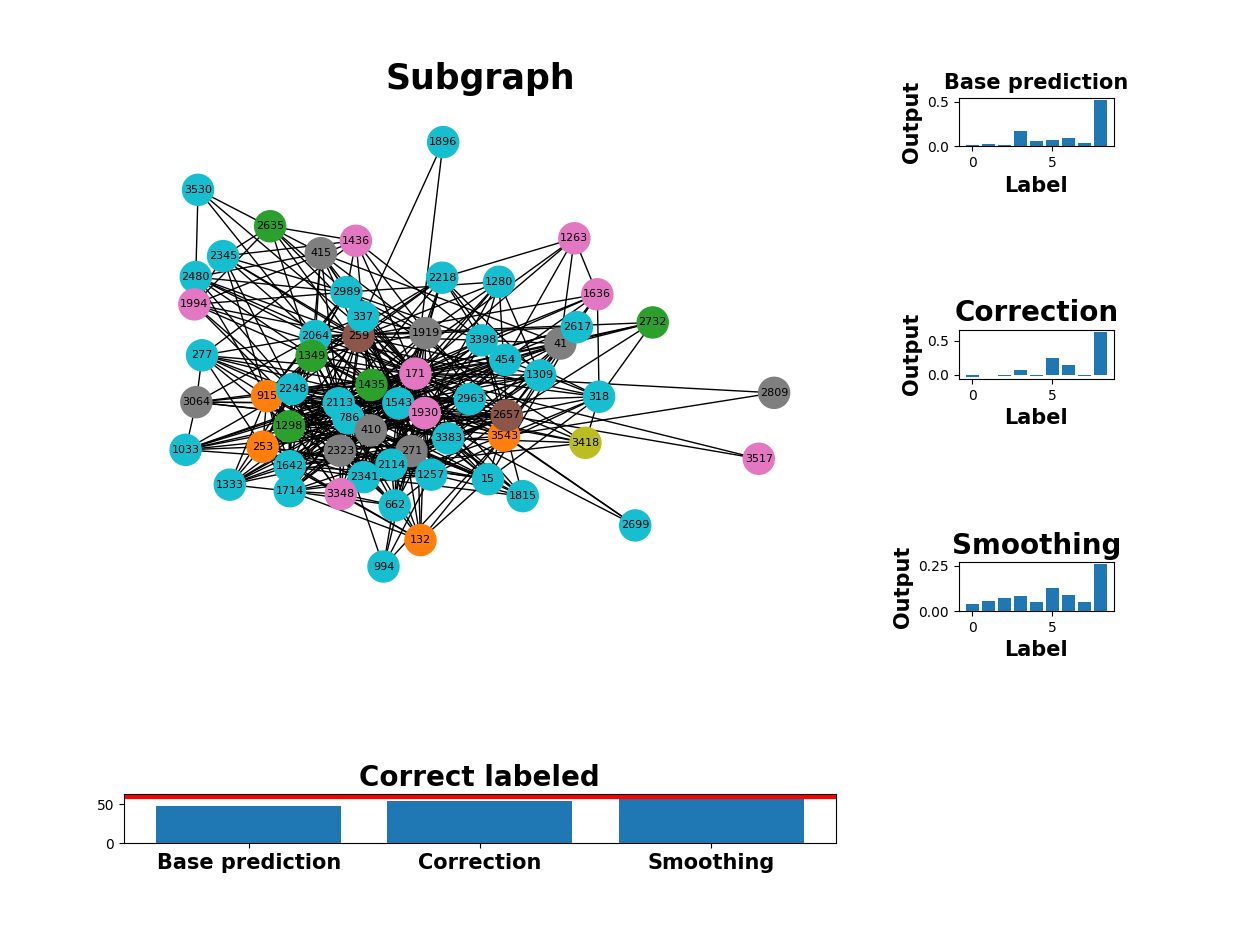}}
    \hfill
   \caption{The subgraph visualization showcases a center node 171 and its one-hop neighbors from Rice31 with 10\% known labels using PL base prediction.}
   \label{fig-rice-171-app} 
\end{figure*}

Visualizations of more subgraphs are shown in Figure \ref{fig-rice-10-app} and \ref{fig-rice-171-app}. Comparison between NLCS and C\&S in Figure \ref{fig-rice-10-app} shows that NLCS makes 27 and 31 correct predictions out of 33 neighbors in correction and smoothing respectively given base prediction's 23 correct predictions, while C\&S makes 26 and 28. On the other hand, a Comparison between NLCS and C\&S in Figure \ref{fig-rice-171-app} shows that NLCS makes 56 and 59 correct predictions out of 60 neighbors in correction and smoothing respectively given base prediction's 48 correct predictions, while C\&S makes 54 and 57. These two figures illustrate the complex relationship in the graph and the challenging situation regarding the variety of neighbor labels, in which NLCS utilizes higher-order representation to improve prediction accuracy. 

\begin{figure}[h]
    \centering
    \subfloat[\label{fig-rice-cs} Rice31]{
    \includegraphics[width=0.48\linewidth]{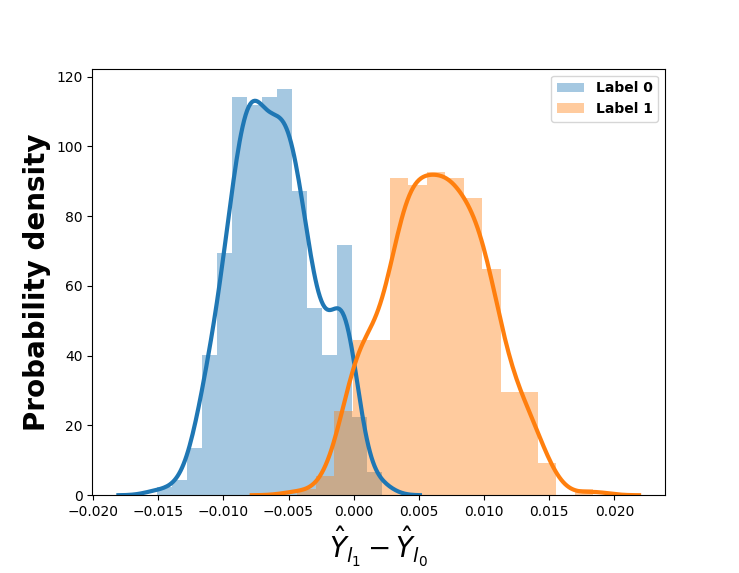}}
    \hfill
    \subfloat[\label{fig-cora-cs} Cora]{
    \includegraphics[width=0.48\linewidth]{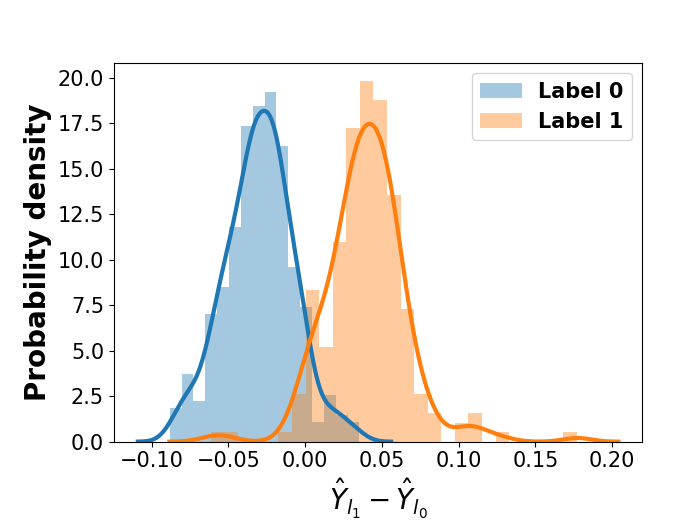}}
    \hfill
   \caption{Demonstrate the distribution of the difference value between the first two output values regarding NLCS for the Rice31 and Cora datasets.}
   \label{fig-cs-app} 
\end{figure}

\begin{figure*}[h]
    \centering
    \begin{minipage}{.48\linewidth}
        \centering
        \subfloat[\label{fig-fout-3pca}]{
        \includegraphics[width=1\linewidth]{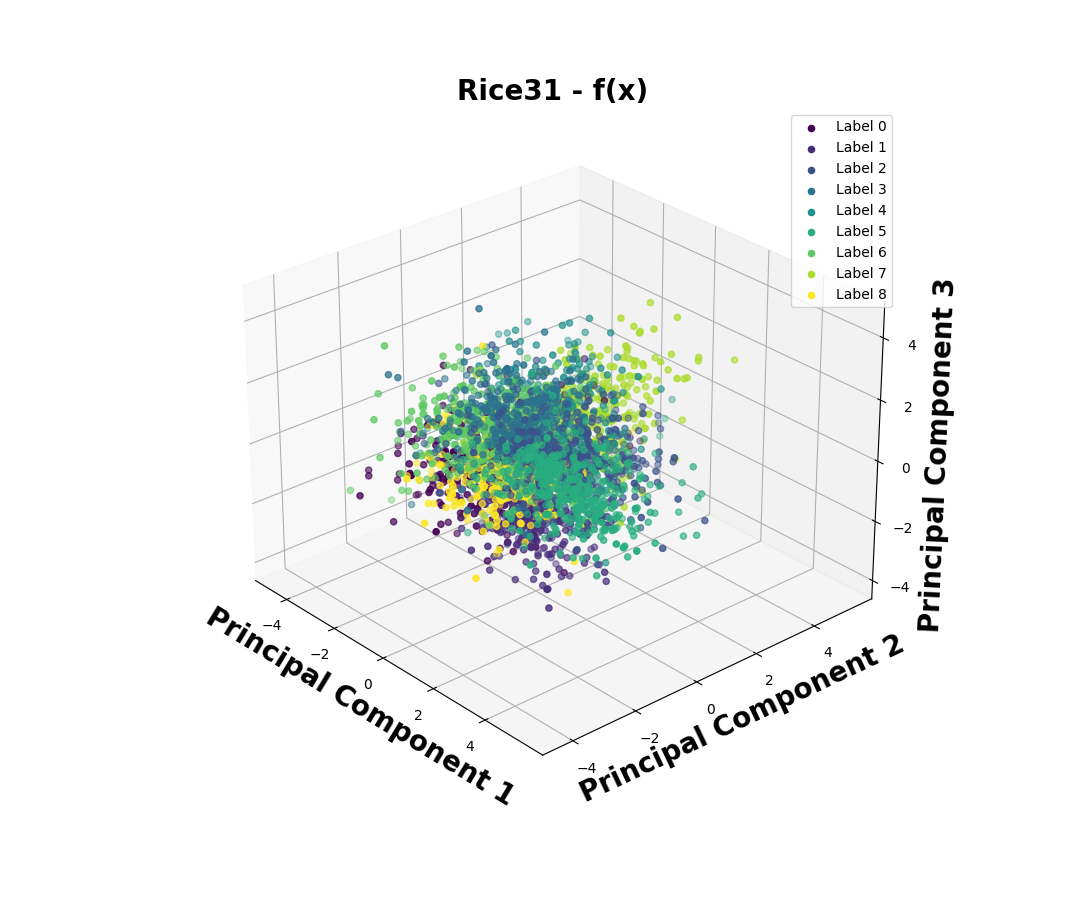}}
    \end{minipage}
    \begin{minipage}{.48\linewidth}
        \centering
        \subfloat[\label{fig-model-3pca}]{
        \includegraphics[width=1\linewidth]{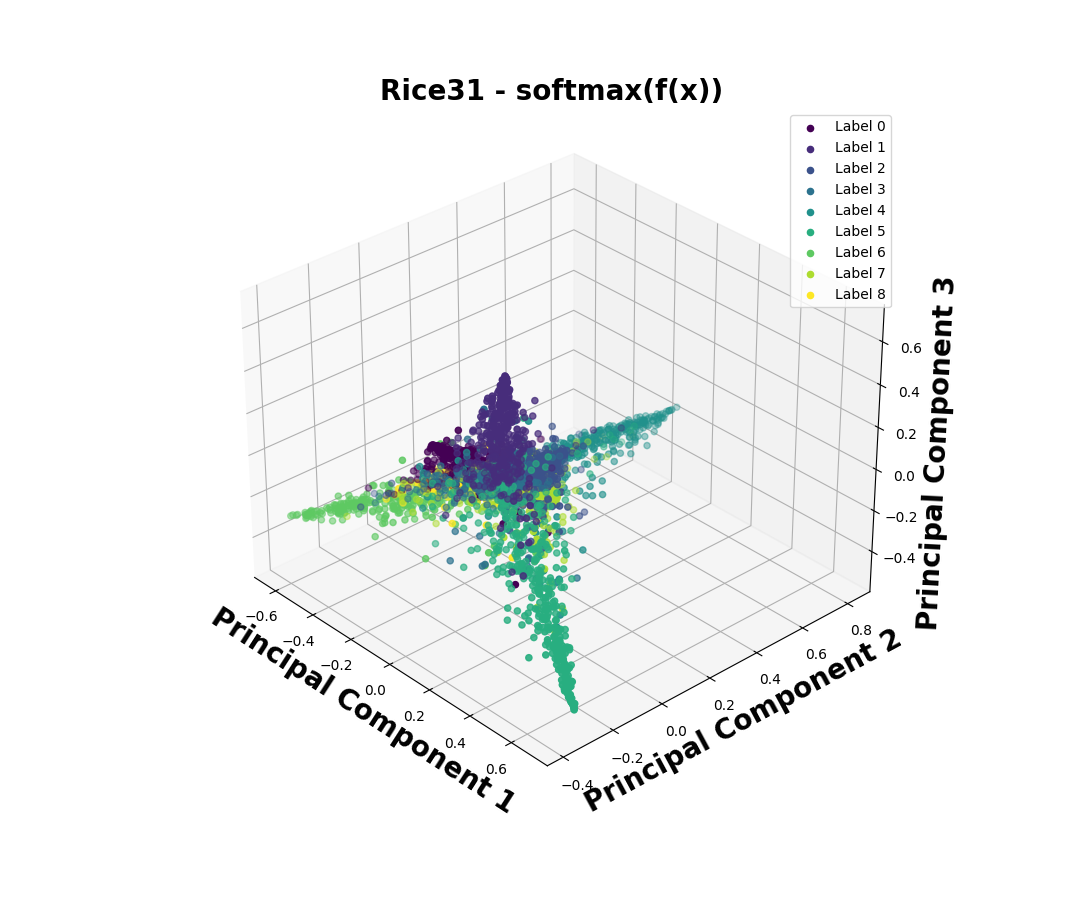}}
    \end{minipage}\par\medskip
    \begin{minipage}{.48\linewidth}
        \centering
        \subfloat[\label{fig-correction-3pca}]{
        \includegraphics[width=1\linewidth]{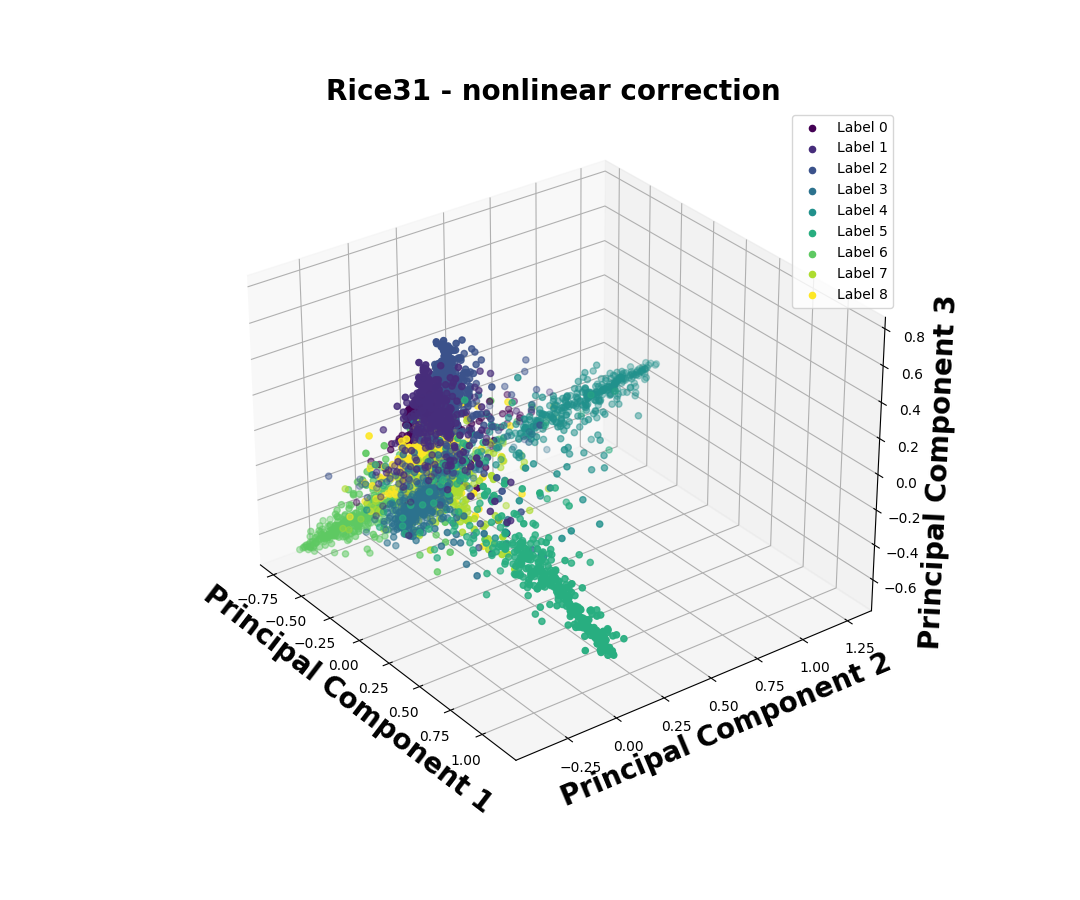}}
    \end{minipage}
    \begin{minipage}{.48\linewidth}
        \centering
        \subfloat[\label{fig-smooth-3pca}]{
        \includegraphics[width=1\linewidth]{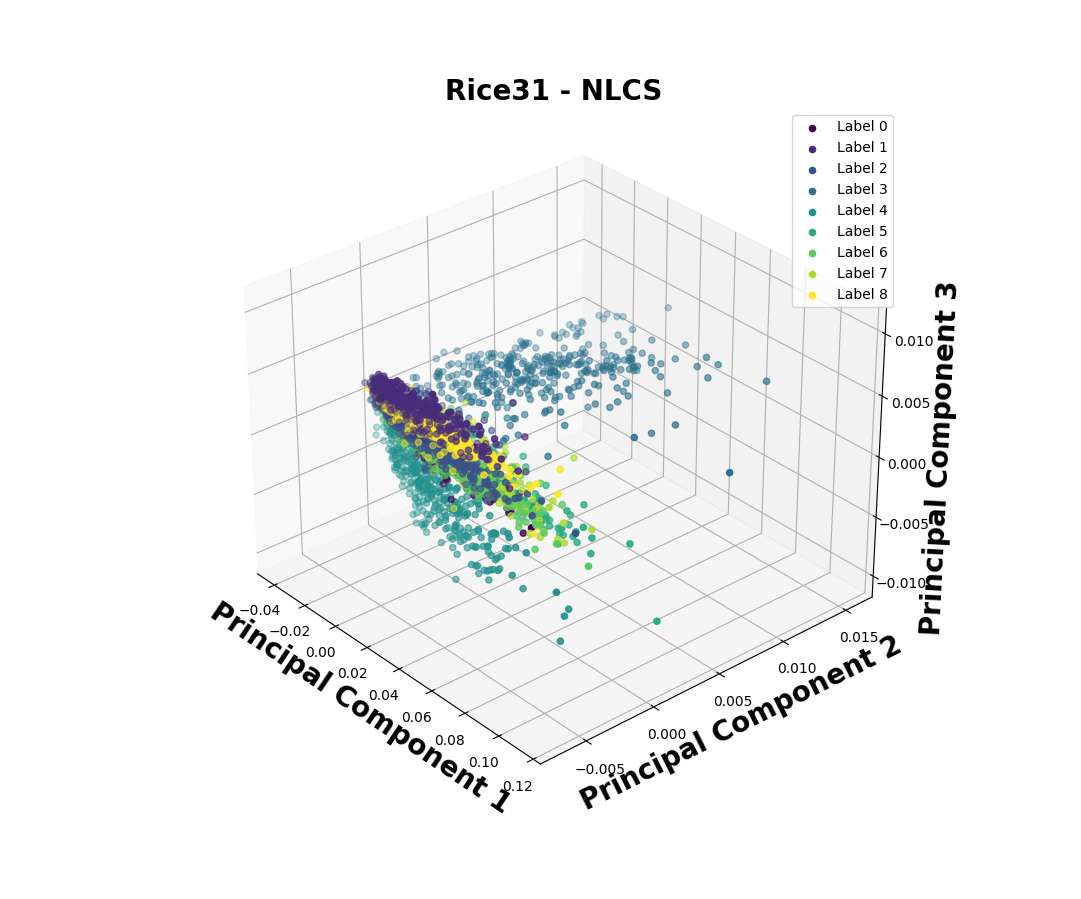}}
    \end{minipage}
    \caption{The PCA visualization on the output value of Rice31 with 5\% known labels. }
    \label{fig-pca-app} 
\end{figure*}

\end{document}